\documentclass[10pt,twocolumn,letterpaper]{article}

\usepackage{cvpr}              % To produce the CAMERA-READY version
\usepackage[dvipsnames]{xcolor}

\usepackage{multirow}
\usepackage{gensymb}
\usepackage{colortbl}
\usepackage{xcolor}
\usepackage{overpic}
\usepackage{dsfont}

\usepackage[accsupp]{axessibility} % Improves PDF readability for those with visual impairments.

\newtheorem{hypo}{Hypothesis}[section]

\definecolor{cvprblue}{rgb}{0.21,0.49,0.74}
\usepackage[pagebackref,breaklinks,colorlinks,citecolor=cvprblue]{hyperref}

\def\paperID{4878} % *** Enter the Paper ID here
\def\confName{CVPR}
\def\confYear{2024}

\title{Extend Your Own Correspondences: Unsupervised Distant Point Cloud Registration by Progressive Distance Extension}

\author{Quan Liu$^{1}$ \quad Hongzi Zhu$^{1}$\thanks{Corresponding author} \quad Zhenxi Wang$^{1}$ \quad Yunsong Zhou$^{1}$ \quad Shan Chang$^{2}$ \quad Minyi Guo$^{1}$\\
\vspace{-0.4cm}\\
{ $^1$Shanghai Jiao Tong University} \quad
{ $^2$Donghua University}\\
\vspace{-0.45cm}\\
{\tt\small \href{https://github.com/liuQuan98/EYOC}{https://github.com/liuQuan98/EYOC}}
}
\begin{document}
\maketitle
\begin{abstract}
Registration of point clouds collected from a pair of distant vehicles provides a comprehensive and accurate 3D view of the driving scenario, which is vital for driving safety related applications, yet existing literature suffers from the expensive pose label acquisition and the deficiency to generalize to new data distributions. 
In this paper, we propose EYOC, an unsupervised distant point cloud registration method that adapts to new point cloud distributions on the fly, requiring no global pose labels. 
The core idea of EYOC is to train a feature extractor in a progressive fashion, where in each round, the feature extractor, trained with near point cloud pairs, can label slightly farther point cloud pairs, enabling self-supervision on such far point cloud pairs. This process continues until the derived extractor can be used to register distant point clouds. 
Particularly, to enable high-fidelity correspondence label generation, we devise an effective spatial filtering scheme to select the most representative correspondences to register a point cloud pair, and then utilize the aligned point clouds to discover more correct correspondences.
Experiments show that EYOC can achieve comparable performance with state-of-the-art supervised methods at a lower training cost. Moreover, it outwits supervised methods regarding generalization performance on new data distributions. 

\end{abstract}    
\section{Introduction}
\label{sec:intro}

Registering point clouds obtained on distant vehicles of 5 meters to 50 meters apart \cite{ijcai2023p134,liu2023density} can greatly benefit a rich set of self-driving vision tasks, ranging from detection \cite{zhang2021emp,yu2022dair,zhu2022vpfnet} and segmentation \cite{simon2019complexer,xu2020squeezesegv3} to birds' eye view (BEV) representation \cite{li2022bevformer,qin2023unifusion} and SLAM \cite{montemerlo2002fastslam,mur2017orb}, and ultimately improve the overall driving safety. Traditional supervised registration methods not only heavily rely on \textit{accurate} pose labels during training \cite{behley2019iccv,Caesar_2020_CVPR,Sun_2020_CVPR} but cannot attain expected performance on new data distributions as they do on existing datasets \cite{choy2019fully,huang2021predator}, making them infeasible to use in real-world driving scenarios. In light of the ever-growing LiDAR-equipped vehicles and the tremendous amount of sequential unlabelled point cloud data, \textit{can we finetune a registration network on a new point cloud distribution with no pose labels so that distant point clouds on the new distribution can be accurately registered on the fly?}

\begin{figure}[t]
  \centering
  \begin{overpic}[scale=0.75]{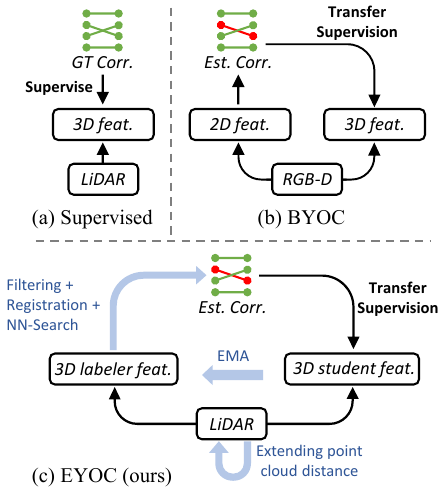}
  \put(71,54.2){\footnotesize\cite{el2021bootstrap}}
  \end{overpic}
  \vspace{-0.2cm}
  \caption{(a) Supervised registration require ground-truth (GT) pose, and (b) BYOC requires RGB-D images for supervision \cite{el2021bootstrap}. (c) In contrast, EYOC acquires supervision from LiDAR sequences directly, enabling single-modal unsupervised training.}
  \label{fig:motivation}
  \vspace{-0.4cm}
\end{figure}

In the literature, a rich set of supervised indoor \cite{huang2021predator,li2022lepard,qin2022geometric,yu2023peal,lee2021deeppro,yew2022regtr} or synthetic \cite{wang2019prnet,yew2020rpm,fu2021robust,ali2021rpsrnet} low-overlap registration methods have been proposed. Most of these methods simply fail on outdoor distant point clouds due to the patch-similarity assumption \cite{lu2021hregnet,yu2023peal} or structural prior such as optimal-transport \cite{yu2021cofinet,qin2022geometric} no longer hold. While simpler networks (\textit{e.g.}, CNNs) showed better robustness on distant point clouds \cite{choy2019fully,huang2021predator}, they still need expensive ground-truth poses for training, as depicted in \cref{fig:motivation}(a).
% \yunzhe{This paragraph seems also related work. Maybe we can move it to 2nd paragraph.}
As pointed out by Banani and Johnson \cite{el2021bootstrap}, unsupervised registration is all about establishing correspondences.
% Supervised methods extract spatial correspondences with aligned point cloud pairs, and use them as positives for contrastive learning (CL), as depicted in \cref{fig:motivation}(a). 
% However, acquiring the global position labels could be prohibitively costly with high-end inertial navigators \cite{}, especially for large-scale road tests. 
% However, they encounter generalization issues on long-tail distributions typical in mass deployment \cite{}, which calls for unsupervised methods.
BYOC \cite{el2021bootstrap}, Unsupervised R\&R \cite{el2021unsupervisedr}, and UDPReg \cite{mei2023unsupervised} have bypassed correspondence acquisition in the indoor setting by borrowing correspondences from RGB channel or GMM matching, as depicted in \cref{fig:motivation}(b), but they suffer from the discrepancy between dense surround images and a sparse point cloud in outdoor settings. As a result, there is no successful solution, to the best of our knowledge, to the unsupervised distant point cloud registration problem.

In this paper, we propose \emph{Extend Your Own Correspondences} (EYOC), a fully unsupervised outdoor distant point cloud registration method \emph{requiring neither pose labels nor any input of other modality}. 
% Based on the intuition that inter-LiDAR distance is proportional with registration difficulty, we transfer the idea of BYOC \cite{} to our setup. 
As depicted in \cref{fig:motivation}(c), our core idea is to adopt a progressive self-labeling scheme to train a feature extractor in multiple rounds. 
Specifically, in each round, a labeler model trained with near point cloud pairs can generate correspondence labels for farther-apart point clouds, which are used to train a student model. 
Particularly, the Siamese labeler-student models are synchronized using the exponential moving average (EMA).  
This process repeats until a full-fledged student model, capable of extracting effective features for distant point cloud registration, is obtained. Two main challenges are encountered in the design of EYOC as follows.

%the distance between two point clouds to be registered until reaching the full distant point cloud registration capability.

% This `magic' extension labelling capability comes from our key insight: Features on far-from-LiDAR points are more transferable than features on close-to-LiDAR points when faced with the next step. Consequently, we could vastly reduce false correspondences simply by discarding close-to-LiDAR points. 

%A curated correspondence filtering pipeline empowers the labeler to generate quality supervision, enabling fast uni-modal unsupervised training with unrivaled generalization capability.

First, it is extremely challenging to prevent the self-labelling process from diverging, given the extreme low-overlap and density-variation of a distant point cloud pair, as witnessed even in supervised training \cite{ijcai2023p134}. To deal with this challenge, we take a gradual learning methodology by breaking the hard learning problem into a series of learning steps with increasing learning difficulties.
Specifically, in the first step, considering the spatial locality of two consecutive frames in a LiDAR point cloud sequence, we assume that two consecutive frames approximately have no transformation, which can be used as supervision to train a basic model. 
After the model first converges to a decent set of weights, we enable the labeler-student self-labelling process and gradually extend the interval of training frames in each learning step. As a result, the student model can converge smoothly.

% Lying at the core of EYOC is a correspondence filtering pipeline that helps the 3D feature extractor generate supervision signals for itself and extend its registration distance, instead of simple Lowe filtering in previous literature \cite{el2021bootstrap}, as depicted in \cref{fig:motivation}(c). The pipeline turned out crucial in enabling correspondence discovery under chaotic 3D features.
% EYOC extends the trick to `lend' correspondence (\textit{i.e.}, supervision) from a weaker 3D labeler model to train a stronger 3D student model on longer-distance pairs, then the student replaces the labeler and the process repeats. The process starts with a barebone model which merely registers next-frame point clouds, and self-improves upon itself until it can register distant point clouds.

Second, it is nontrivial for the labeler in one learning step to generate sufficient correspondence labels of high quality for the next harder learning step. We observe \emph{the near-far diversity phenomenon} of LiDAR point clouds, \emph{i.e.}, when the observation distance changes, the point density variation of near objects is larger than that of far objects. This means that features extracted from low-density (far-from-LiDAR) regions are more stable along with distance changes. 
Inspired by this insight, we develop a spatial filtering technique to effectively discover a set of initial quality correspondences in low-density regions.
Furthermore, to obtain more widespread correspondences, we perform a live registration using the initial correspondences followed by another round of nearest-neighbor search (NN-Search) to further dig out and amplify correct correspondences, readied for supervision of the student.

We evaluate EYOC design with trace-driven experiments on three major self-driving datasets, \emph{i.e.}, KITTI \cite{Geiger2012CVPR}, nuScenes \cite{Caesar_2020_CVPR}, and WOD \cite{Sun_2020_CVPR}. EYOC reaps comparable performance with state-of-the-art (SOTA) fully supervised registration methods while outwitting them by $17.4\%$ mean registration recall in an out-of-domain unlabelled setting.
% Thanks to the dataset extension strategy, EYOC also halves the training time compared with the traditional two-step strategy \cite{ijcai2023p134,liu2023density}.
To summarize, our contributions are listed as follows:

\begin{itemize}
    \item We analyzed the \textit{near-far diversity} of point clouds, where low-density regions of a point cloud produce consistent feature correspondences during a distance extension step.
    \item We propose an unsupervised distant point cloud registration method that can effectively adapt to new data distributions without pose labels or other input modalities.
    % \item We propose a data selection strategy where V2V distance grows with training epochs, enabling fast and guaranteed model convergence.
    % \item We propose a spatial filtering technique based on the fact that far-from-LiDAR correspondences exhibit better similarity under extended V2V distance.
    % \item Exploiting the duality between correspondence and pose, we further propose to use registration \& NN-search to amplify the correct correspondences to a trainable level.
    \item The performance and applicability of EYOC are validated with extensive experiments on three self-driving datasets.
\end{itemize}

% However, As geometric feature extractors extract correspondences from pose-aligned point clouds, BYOC transfer RGB correspondences to the depth channel for supervision. Based on the intuition that close point clouds are easier to register than distant point clouds, we transfer that idea to our setup: We train a weak model trained with close-range point clouds, and extend its correspondences to longer-distance point cloud pairs.

% As pointed out by Banani and Johnson \cite{}, the core of unsupervised registration is all about establishing correspondences. While supervised methods extract correspondences from pose-aligned point clouds, unsupervised methods seek help from RGB correspondences (supervision) which are then tranferred to the depth channel: We see that an easy problem (RGB matching) helps a hard problem (3D matching). Based on the intuition that close point clouds are easier to register than distant point clouds, we transfer that idea to our setup as follows. We split the

\section{Related Work}

\subsection{Supervised Registration}
Recent registration techniques are highly monopolized by learning based methods \cite{zeng20173dmatch,deng2018ppfnet,gojcic2019perfect,poiesi2021distinctive,ao2021spinnet,ao2023buffer,choy2019fully,bai2020d3feat,huang2021predator,ijcai2023p134,liu2023density,lu2021hregnet,yu2021cofinet,qin2022geometric,lee2021deeppro,yew2022regtr}, due to both superior accuracy and faster inference speed compared with traditional extractors \cite{johnson1999using,rusu2009fast,tombari2010unique} or pose estimators such as RANSAC \cite{fischler1981random}.

\paragraph{Local feature extractors.} 
Correspondence-based local feature extractors have long been diverged into patch-based methods \cite{zeng20173dmatch,deng2018ppfnet,gojcic2019perfect,poiesi2021distinctive,ao2021spinnet} and fully-convolutional methods \cite{choy2019fully,bai2020d3feat,huang2021predator,ijcai2023p134,liu2023density}.
% where the former reaps better generalization capability and the latter enjoys real-time inference. 
3DMatch \cite{zeng20173dmatch} initiated the patch-based genre, while PointNet \cite{qi2017pointnet}, smoothed density value and reconstruction were later introduced by PPF-Net \cite{deng2018ppfnet}, PerfectMatch \cite{gojcic2019perfect}, and DIP \cite{poiesi2021distinctive}, respectively. The recent pinnacle SpinNet \cite{ao2021spinnet} and BUFFER \cite{ao2023buffer} combine SO(2) equivalent cylindrical features with fully convolutional backbones. On the other hand, following FCGF \cite{choy2019fully}, fully convolutional methods process the point cloud as a whole. KPConv \cite{thomas2019kpconv} backbones are equipped with keypoint detection in D3Feat \cite{bai2020d3feat} and overlap attention in Predator \cite{huang2021predator}. APR \cite{ijcai2023p134} and GCL \cite{liu2023density} further enhanced outdoor distant low-overlap registration with reconstruction and group-wise contrastive learning. We build our method upon fully convolutional methods because they are deemed most suitable for fast and generalizable outdoor registration.

\begin{figure*}[t]
  \centering
  \includegraphics[width=0.8\linewidth]{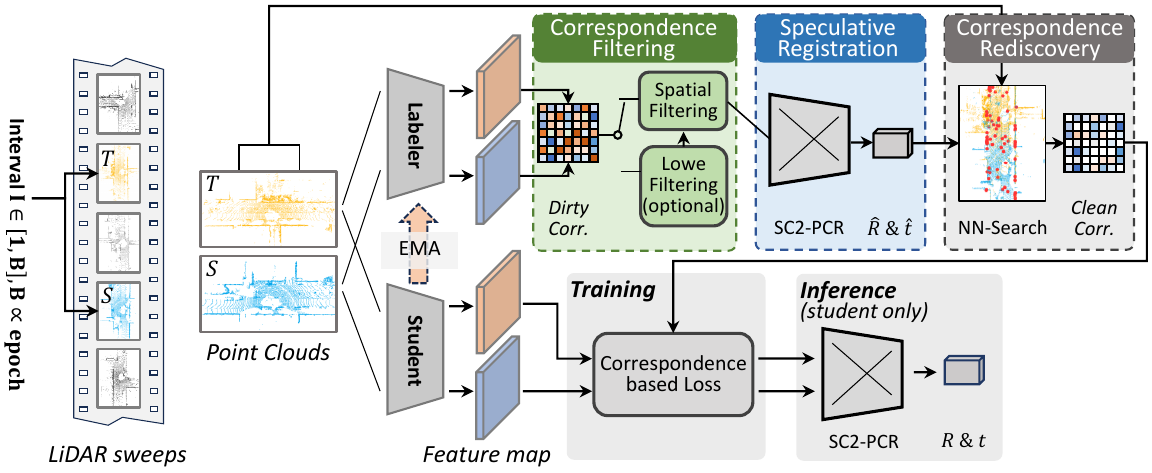}
  \vspace{-0.2cm}
  \caption{\textbf{Overview of Extend Your Own Correspondences (EYOC)}. It exhibits a two-branch student-labeler structure with periodic synchronization, where the labeler generates correspondences for the student. Point cloud pairs are selected at random frame interval, whose range extends with time. Labeler dirty correspondences are filtered before the speculative registration which outputs an estimated pose. Finally, correspondence rediscovery with NN-search on re-aligned input point clouds recovers clean correspondence labels.
%   \yunzhe{What's the output of "Correspondence Rediscovery"? This should be shown in the figure.}
  }
  \label{fig:architecture}
  \vspace{-0.4cm}
\end{figure*}

\paragraph{Pose estimators.} Pose estimators \cite{fischler1981random,zhou2016fast,choy2020deep,bai2021pointdsc,lee2021deep,chen2022sc2,zhang20233d} take in feature maps and output the most probable pose estimation, where RANSAC \cite{fischler1981random} is a common time-consuming baseline. While DGR \cite{choy2020deep}, PointDSC \cite{bai2021pointdsc} and DHVR \cite{lee2021deep} opted for learned correspondence weight with FCNs, Non-local Module, and Hough Voting, respectively, non-parametric methods such as SC$^2$-PCR \cite{chen2022sc2} and MAC \cite{zhang20233d} hit higher marks through the second order compatibility or maximal clique search.

\paragraph{Keypoint-free registration.} Keypoint-free methods borrowed the idea of superpixels \cite{giraud2017superpatchmatch} from image matching to match heavily downsampled points (\textit{i.e.}, superpoints), each representing a local patch \cite{lu2021hregnet,yu2021cofinet,qin2022geometric,yu2023peal,lee2021deeppro,yew2022regtr}. HRegNet \cite{lu2021hregnet} proposed to refine global pose with different stages of downsampling. CoFiNet \cite{yu2021cofinet}, GeoTransformer \cite{qin2022geometric}, and PEAL \cite{yu2023peal} treat superpoints as seeds and match promising seed patches only. Another line of work, DeepPRO \cite{lee2021deeppro} and REGTR \cite{yew2022regtr}, regress correspondences directly without feature matching. However, their assumption that superpoint patches should share high overlap no longer holds considering extreme density-variation and low-overlap.
% Consequently, most of them fail to converge on distant point clouds.

\subsection{Unsupervised Registration}

Compared with supervised methods, unsupervised registration is less explored especially for the outdoor scenario. BYOC \cite{el2021bootstrap} highlighted that random 2D CNNs could generate image correspondences good enough to supervise a 3D network, therefore indoor RGB-D images are used for self-supervision. UnsuperisedR\&R \cite{el2021unsupervisedr} in turn seeked help from differentiable rendering of RGB-D images as mutual supervision after differentiable registration. UDPReg \cite{mei2023unsupervised} enforced multiple losses on GMM matching to generate correspondences for indoor point clouds. However, outdoor unsupervised registration remain an exciting yet unexplored field of research, calling for more work on this area.

\section{Problem Definition}
% \yunzhe{This part should be an individual section? What's the different between distant registration and other registration?}
Given two point clouds $\mathcal{S}\in \mathbb{R}^{n\times 3}, \mathcal{T}\in \mathbb{R}^{n\times 3}$, point cloud registration aims to uncover their relative transformation $R\in SO(3), t\in \mathbb{R}^{3}$ so that $\mathcal{S}R^T+t^T$ aligns with $\mathcal{T}$. When the LiDARs are placed on two distant vehicles separated at a distance of $d\in [5m,50m]$, the sub-problem is referred to as \textit{distant} point cloud registration \cite{ijcai2023p134}. Contrary to previous settings \cite{huang2021predator,lee2021deeppro,yew2022regtr}, distant point clouds share extreme low-overlap and density-variation leading to network divergence when directly applied to training. This is usually mitigated through a staged training strategy with pretraining on high-overlap pairs and finetuning on low-overlap pairs \cite{ijcai2023p134}.
% This is often done through establishing a set of correspondences through feature matching and training them using contrastive loss \cite{zeng20173dmatch,choy2019fully}. With a set of feature correspondences, the transformation can be solved with RANSAC-like algorithms. On the other hand, once a transformation has been determined, one can also extract correspondences by performing nearest-neighbor search between point clouds.
\section{Method}

% \subsection{Overview}
\label{sec:overview}
The overview of EYOC is illustrated in \cref{fig:architecture}, which composes of a siamese student-labeler network structure followed by correspondence filtering, speculative registration, and correspondence rediscovery.
During training, two distant point clouds, $\mathcal{S}, \mathcal{T}$, are fed into the student and labeler networks to extract point-wise features $F_\mathcal{S}^{stu}, F_\mathcal{S}^{lab}\in \mathbb{R}^{n\times k}$ and $F_\mathcal{T}^{stu}, F_\mathcal{T}^{lab}\in \mathbb{R}^{m\times k}$. The labeler features are then processed by correspondence filtering to obtain a decent correspondence set $C^{lab}=\{(i,j)|p_i\in \mathcal{S}, q_j\in \mathcal{T}\}$. It is later fed into speculative registration to decide an optimal transformation $\hat{R}\in SO(3), \hat{t}\in \mathbb{R}^{3}$ between $\mathcal{S}$ and $\mathcal{T}$.
% , along with explicit failure detection which triggers to skip training on misaligned pairs. 
The high-fidelity estimated transformation is used to re-align input point clouds, which allows us to rediscover correspondences using NN-Search for supervision of the student.

\subsection{Extension of Point Cloud Distance}
Unlike the supervised setting, it is impossible to calculate the accurate distance between LiDARs in the unsupervised setting. However, leveraging the spatial locality of LiDAR sequences, we can limit the translational upper bound by limiting the frame interval $I$ between two frames in a sequence. Improving upon the staged training strategy \cite{ijcai2023p134}, we propose to randomly select the frame interval $I\in \mathbb{N}^+, I\in[1,B]$ for every pair, where $B$ grows from 1 to 30 during the course of training, forming 30 tiny steps. When $B=1$, we assume identity transformation and apply supervised training. Our \textit{progressive distance extension} strategy increases the problem difficulty gradually to facilitate smooth convergence.

\subsection{Labeler-Student Feature Extraction}
Given a pair of distant point clouds, we pass them through two homogeneous 3D sparse convolutional backbones parameterized by $W^{lab}$ and $W^{stu}$, to obtain point-wise feature maps. The student is periodically updated to the labeler in a gradual manner of exponential moving average (EMA), which keeps the labeler both stable and up-to-date, facilitating consistent label generation. Specifically, we update the labeler weights as in \cref{eq:ema} after every epoch, where $\lambda\in[0,1)$ is a decay factor:

\begin{equation}
    \label{eq:ema}
    W_{t+1}^{lab} \longleftarrow \lambda W_t^{lab} + (1-\lambda) W_t^{stu}
\end{equation}

\begin{figure}[t]
  \centering
%   \vspace{-0.2cm}
%   \includegraphics[width=0.5\linewidth]{figure/visual_similarity_1.png}\includegraphics[width=0.5\linewidth]{figure/visual_similarity_3.png}
  \includegraphics[width=0.4\linewidth]{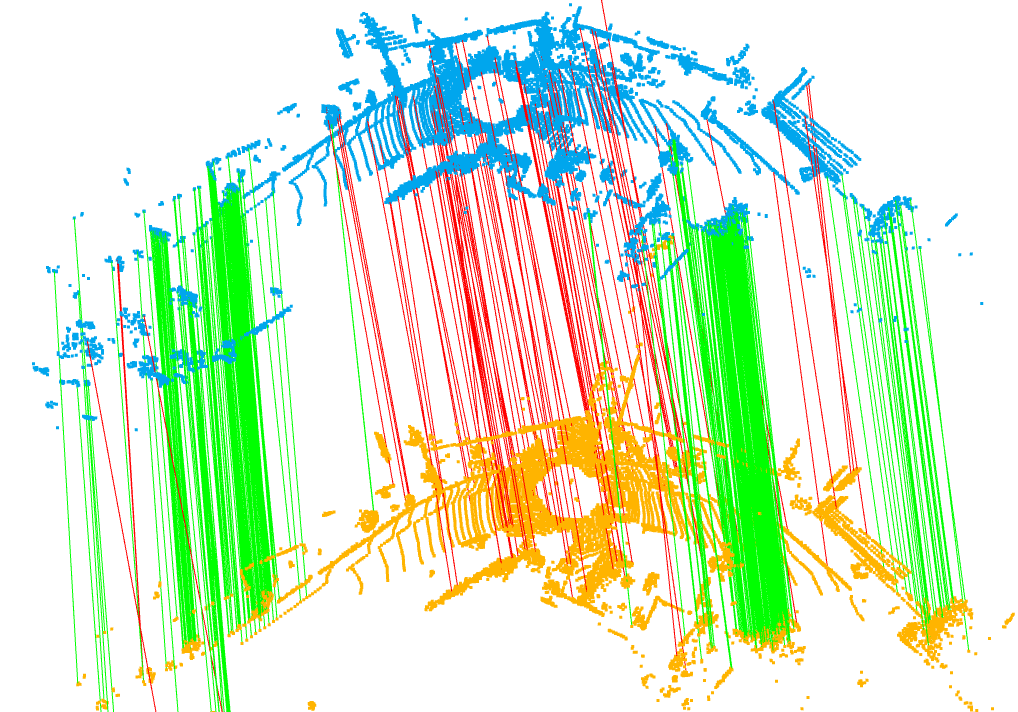}\includegraphics[width=0.6\linewidth]{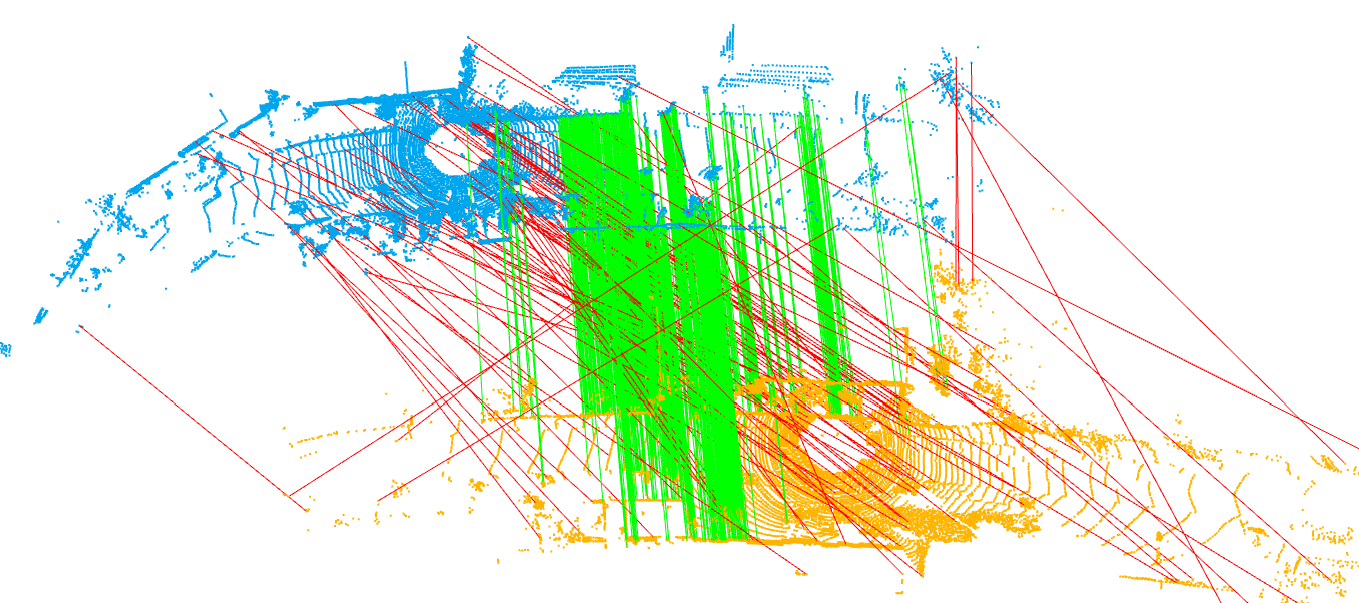}
  \vspace{-0.4cm}
  \caption{\textbf{The dirty correspondence labels generated by closer-range labeler (Left: $B=1$; Right: $B=10$) on farther-apart point clouds (Left: $d=10m$; Right: $d=30m$) in KITTI \cite{Geiger2012CVPR} before spatial filter.} Correct ones are colored green and false ones red. Close-to-LiDAR features are less generalizable to farther pairs than far-from-LiDAR features.}
  \label{fig:visual_similarity}
  \vspace{-0.5cm}
\end{figure}

\subsection{Correspondence Filtering}
The correspondence filtering module aims to maximize the portion of correct correspondences produced by the labeler to enable unsupervised label generation. Different from BYOC, random 3D CNNs cast much worse correspondences than random 2D CNNs \cite{ulyanov2018deep,rosenfeld2019intriguing,el2021bootstrap}, so the dirty correspondences obtained by matching 3D labeler features $F_\mathcal{S}^{lab}$ and $F_\mathcal{T}^{lab}$ is likely to be rife with different fault patterns from RGB-D images. With that in mind, we investigate two types of filtering techniques on both feature space and Euclidean space based on data-centric observations.

\paragraph{Lowe filtering.} Previous literature \cite{el2021unsupervisedr, el2021bootstrap} have found Lowe's Ratio \cite{longuet1981computer} a good match for rating the most unique correspondences on indoor RGB-D point clouds. Specifically, given two corresponding features $f_\mathcal{S}^i \in F_\mathcal{S}^{lab}, f_\mathcal{T}^j\in F_\mathcal{T}^{lab}$, the significance is calculated according to \cref{eq:lowe}, where $D(\cdot,\cdot)$ denotes the cosine similarity. Contrary to previous literature, we find Lowe filtering to deteriorate correspondence quality drastically as discussed in \cref{sec:ablation}.

\begin{equation}
    \label{eq:lowe}
    \omega_{i,j} = 1 - \frac{D(f_\mathcal{S}^i,f_\mathcal{T}^j)}{\min_{f_\mathcal{T}^k\in F_\mathcal{T}^{lab}, k\neq j}{D(f_\mathcal{S}^i,f_\mathcal{T}^k)}}
\end{equation}

% The top 5000 correspondences are selected. Contrary to previous literature, we find Lowe filtering to deteriorate correspondence quality drastically as discussed in \cref{sec:ablation}. This, in turn, hints that correct feature pairs are close competitors with false ones in feature space, making metric-wise filtering rather difficult.

\paragraph{Spatial characteristic of labeler correspondences.} 
In response to the failure of Lowe filtering, we conduct a label-driven investigation based on the \textit{the near-far diversity phenomenon}, where far objects should have more consistent densities when the viewpoint undergoes displacements. We hereby examine the quality of raw feature correspondences for a labeler model on farther-apart point clouds than those in the labeler's training set, as depicted in \cref{fig:visual_similarity}, and propose the following hypothesis: 

\begin{hypo}
    Correct correspondences are more likely to be clustered in low-density regions far from the LiDARs during the distance extension.
    \label{hyp:corr}
\end{hypo}
\label{sec:hypothesis}

\begin{figure}[t]
  \centering
%   \vspace{-0.2cm}
%   \includegraphics[width=\linewidth]{figure/analysis.pdf}
  \includegraphics[width=\linewidth]{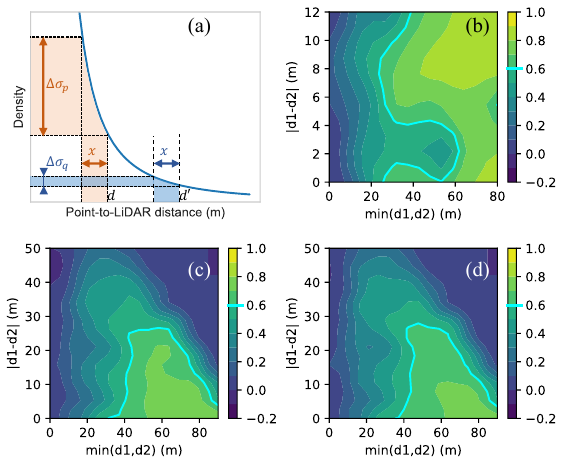}
  \vspace{-0.7cm}
  \caption{\textbf{Visual groundings} for our hypothesis on KITTI \cite{Geiger2012CVPR}. (a) Density of close-to-LiDAR points are more sensitive to movement than far-from-LiDAR points. (b-d) Cosine similarity of correspondences with its distance to two LiDARs, $d_1, d_2$, under (b) $I\in [1,1]$, (c) $I\in [1,15]$, and (d) $I\in [1,30]$.}
  \label{fig:theoretical_similarity}
  \vspace{-0.5cm}
\end{figure}

\noindent\textit{Proof.} We provide the rationale of a simplified case here based on the LiDAR sensor model \cite{lawin2018density}.
A LiDAR can be modeled as a light source emitting light uniformly in all directions, and the probability density of a point being scanned is proportional to its energy absorption rate.
Specifically, given two points in the world coordinate $p=(d,0,0)^T, q=(d',0,0)^T, 0<d<d'$ and the current LiDAR center $O=(0,0,0)^T$, their respective densities are $\sigma_p=\frac{\alpha}{d^2}, \sigma_q=\frac{\alpha}{{d'}^2}$, where $\alpha$ is an unknown constant depending on the LiDAR resolution and incident angle, which we assume are the same for $p$ and $q$. Suppose the LiDAR center now moves to $O'=(x,0,0)^T$ where $0<x<d<d'$, the delta densities are $\Delta\sigma_p=\frac{\alpha}{(d-x)^2}-\frac{\alpha}{d^2}, \Delta\sigma_q=\frac{\alpha}{(d'-x)^2}-\frac{\alpha}{{d'}^2}$. It is easy to prove that $\Delta\sigma_p > \Delta\sigma_q$, as illustrated in \cref{fig:theoretical_similarity}(a). As widely acknowledged, CNN features are sensitive to density variation \cite{wu2019pointconv,te2018rgcnn,yew20183dfeat,huang2021predator,qin2022geometric}, therefore making close-range point features less robust under vehicle translation.

% \yunzhe{This part should be put above and emphasized.}
\paragraph{Spatial filtering design.}
Based on the findings, we quantitatively explore the relationship between distance from a correspondence to two LiDARs denoted by $d_1, d_2$, and the cosine similarity of that feature correspondence, as depicted in \cref{fig:theoretical_similarity}(b-d). 
% The result is averaged out during one epoch. 
We refer readers to Appendix \cref{sec:appendix_other_datasets} for similar results on other datasets. We confirm that close-to-LiDAR regions contain most correspondences, but are consistently under-performing and, therefore, could be purged to improve supervision quality.
% \yunzhe{Maybe reviewers would like to know why.} 
We hereby propose two sets of spatial filtering strategies: 
\begin{itemize}
    \item \textbf{Hard}: Discard points where $\min(d_1,d_2)<d_{thresh}$, regardless of training progression;
    \item \textbf{Adaptive}: Discard regions with $\leq s_{thresh}$ similarity in \cref{fig:theoretical_similarity}, where the decision boundary at $s_{thresh}=0.6$ is highlighted in cyan.
    The similarities are exhaustively recorded from the pretraining dataset.
\end{itemize}

Emperically, both methods suffice to cut over 70\% of the false correspondences while only 9\% correct correspondences are discarded. 

\subsection{Speculative Registration}

After the correspondence filtering, we adopt a SOTA registration algorithm SC$^2$-PCR \cite{chen2022sc2} for accurate real-time registration to amplify the most promising set of correspondences. Although the correspondences have been heavily cleansed down to several hundred pairs, only 20\% among which are correct on average, which is below the bar for direct supervision as discussed in \cref{sec:ablation}; However, literature has shown that this ratio is high enough for a successful registration \cite{chen2022sc2,zhang20233d}. Intuitively, if the input point clouds could be correctly registered, we could imitate fully-supervised training where correspondences are obtained directly from aligned input point clouds instead of matched features. Moreover, the searched nearest neighbors can vastly outnumber the heavily filtered labeler correspondences, making the training process more data-efficient.
Therefore, we propose to obtain an estimated pose $\hat{R}\in SO(3), \hat{t}\in \mathbb{R}^3$ between the input point clouds $\mathcal{S},\mathcal{T}$ on the fly with real-time registration algorithms.

% \paragraph{Registration failure detection.} Another sweet finding is that we could estimate the chances of successfully registering the pair, simply through looking at an intermediate value of SC$^2$-PCR, called $seedwise\_fitness$. 

\subsection{Correspondence Rediscovery}

With the input point clouds $S,T$ and the estimated transformation $\hat{R}\in SO(3), \hat{t}\in \mathbb{R}^3$, we could simply follow supervised training to search correspondences for dense supervision. Specifically, we transform $\mathcal{S}'=R^T \mathcal{S}+t^T$, and obtain the nearest neighbors according to \cref{eq:corr}, where $\beta_{inlier}=2m$ is a loosened match threshold to tolerate minor pose errors.
\vspace{-0.1cm}
\begin{equation}
    \label{eq:corr}
    \begin{split}
        C_{ST}=\bigg\{(i,j)\bigg|p_\mathcal{S}^i\in \mathcal{S}', j=\mathop{\arg\min}\limits_{p_\mathcal{T}^j\in \mathcal{T}}{||p_\mathcal{S}^i-p_\mathcal{T}^j||},\textcolor{white}{\bigg\}} \\
        \textcolor{white}{\bigg\{}s.t.\ ||p_\mathcal{S'}^i-p_\mathcal{T}^j||<\beta_{inlier}\bigg\}
    \end{split}
\end{equation}

\subsection{Loss Design}

We adopt the widely-used Hardest-Contrastive Loss \cite{choy2019fully} as the training loss for the student. As nearest-neighbor search is not differentiable, we only back-propagate gradients to the student but not the labeler. Specifically, the loss is formulated as \cref{eq:loss}:\\
\begin{equation}
    \label{eq:loss}
    \small
    \begin{split}
        L=\frac{1}{|C_{\mathcal{S}\mathcal{T}}|}\sum_{(i,j)\in C_{\mathcal{S}\mathcal{T}}}{\left[m + P(f_\mathcal{S}^i,f_\mathcal{T}^j) - \min_{j\neq k\in\mathcal{N}}{P(f_\mathcal{S}^i,f_\mathcal{T}^k)}\right]_+} \\
        +\frac{1}{|C_{\mathcal{T}\mathcal{S}}|}\sum_{(j,i)\in C_{\mathcal{T}\mathcal{S}}}{\left[m + P(f_\mathcal{T}^j,f_\mathcal{S}^i) - 
        \min_{i\neq k\in\mathcal{N}}{P(f_\mathcal{T}^j,f_\mathcal{S}^k)}\right]_+}
    \end{split}
\end{equation}
\vspace{-0.2cm}

Where $\mathcal{N}$ is a subset of feature indices, $m$ is the positive margin, $[\cdot]_+$ rounds negative values to 0, $P(\cdot, \cdot)$ denotes the squared distance between two vectors. $C_{\mathcal{T}\mathcal{S}}$ follows \cref{eq:corr} but is calculated in the reverse direction from $\mathcal{T}$ to $\mathcal{S}$.

\section{Results}

We demonstrate the superiority of EYOC against state-of-the-art methods on three major self-driving datasets, KITTI \cite{Geiger2012CVPR}, nuScenes \cite{Caesar_2020_CVPR}, and WOD \cite{Sun_2020_CVPR}.
% We aim to prove that EYOC can be a cheaper solution to the traditional labelling-training pipeline, having comparable performance with supervised training but incurring negligible additional cost.
We then provide an ablation study, finetuning strategies, and time analysis. Visualizations for the labeler are available in \cref{fig:visualization}.

\subsection{Experiment Setup}

\paragraph{Datasets.} Apart from our progressive dataset extension strategy, shorthanded as \textit{progressive dataset}, we also follow existing literature \cite{ijcai2023p134,liu2023density} to prepare the point cloud pairs based on the distance between two LiDARs, referred to as \textit{traditional dataset}. The latter works under supervised settings, where the point cloud pairs have a random Euclidean distance between two LiDARs, denoted with $d\in [M,N]$ in meters. The traditional datasets are also used during all test sections. On the other hand, progressive datasets work for either supervised or unsupervised training, where point cloud pairs are selected with a random frame interval $I\in [1,B]$ due to the absence of pose labels. We set the initial value to $B=1$ which grows linearly to $B=30$ during 200 epochs. All datasets are cut into train-val-test splits by official recommendations.

% \paragraph{Purpose of comparison.} Given a pretrained model and a new out-of-distribution dataset, unsupervised methods can perform finetuning while supervised methods cannot, causing a performance increase. We also prove that EYOC can be a cheaper solution to the traditional labelling-training pipeline, having comparable performance with supervised training but incurring negligible additional cost.

\begin{table*}[t]
  \centering
  \small
  \resizebox{\linewidth}{!}{
  \begin{tabular}{@{}cclcccccccccc@{}}
    \toprule
    % \hline
    % \hline
    Test &\multirow{2}{*}{No.}& \multirow{2}{*}{Method}  & Pretrain & Finetune & \multirow{2}{*}{Supervised} & Progressive & \multirow{2}{*}{mRR}& \multicolumn{5}{c}{RR @ $d \in$}\\
    \cline{9-13}
    Set & & & Dataset & Dataset &  & Dataset & &[5,10]	&[10,20] &[20,30] &[30,40] &[40,50]\\
    % \hline
    % 
    % 
    % KITTI
    % 
    % 
    % \hline
    % \cmidrule{2-13}
    \midrule
        \multirow{12}{*}{\rotatebox{90}{KITTI}} &\multirow{2}{*}{a} &FCGF \cite{choy2019fully}	      &WOD &-  & \checkmark &-  &71.8 	&98.0 	&92.5 	&85.0 	&52.6 	&30.7  \\
        % &&FCGF + C	     &WOD &-  &\checkmark & \\
    &&Predator \cite{huang2021predator}	 &WOD &-  &\checkmark & -  &72.3 	&99.5 	&\underline{98.9}	&\underline{\underline{90.9}} 	&56.8 	&15.3 \\
    % &SpinNet          \cite{ao2021spinnet}	     &WOD & - & -       &35.6      &97.6	&73.1	&7.3	&0.0	&0.0\\
    % &D3Feat           \cite{bai2020d3feat}	      &WOD & - & -      &52.5     &98.7	&86.8	&52.7	&20.0   &4.5\\
    % &CoFiNet          \cite{yu2021cofinet}	      &WOD & -& -       &68.6    &{99.6} 	&94.2 	&80.0 	&44.8 	&24.3\\
    % &GeoTransformer   \cite{qin2022geometric}	&WOD & - & - &39.0    &97.9 	&88.3 	&8.3 	&0.7 	&0.0\\
    \cmidrule{2-13}
    &\multirow{7}{*}{b} &FCGF \cite{choy2019fully}	          &- & KITTI &\checkmark & - &77.4 	&98.4 	&95.3 	&86.8 	&69.7 	&36.9 \\
    &&FCGF + C	      &- & KITTI &\checkmark & \checkmark &\underline{84.6} 	&\textbf{100.0} 	&97.5 	&90.1 	&\underline{79.1} 	&\underline{56.3}  \\
    &&Predator \cite{huang2021predator}	  &- & KITTI &\checkmark & -  &\textbf{87.9} 	&\textbf{100.0} 	&\underline{\underline{98.6}} 	&\textbf{97.1} 	&\textbf{80.6} 	&\textbf{63.1} \\
    % &BUFFER       &&- &KITTI & -  &&&&&&\\
    &&SpinNet* \cite{ao2021spinnet}	     &-&KITTI &\checkmark & -  &39.1 	&99.1 	&82.5 	&13.7 	&0.0 	&0.0\\
    &&D3Feat* \cite{bai2020d3feat}	     & -&KITTI &\checkmark & -  &66.4 	&\underline{\underline{99.8}} 	&98.2 	&90.7 	&38.6 	&4.5\\
    &&CoFiNet \cite{yu2021cofinet}	      &-&KITTI &\checkmark & -  &82.1 	&\underline{99.9} 	&\textbf{99.1} 	&\underline{94.1} 	&\underline{\underline{78.6}} 	&38.7\\
    &&GeoTrans.* \cite{qin2022geometric}	&-&KITTI &\checkmark & - &42.2 	&\textbf{100.0} 	&93.9 	&16.6 	&0.7 &0.0\\
    % &&GCL-Conv \cite{liu2023density}	&-&KITTI & - &93.5 &99.0 	&98.8 	&96.1 	&91.7 	&82.0 \\
    \cmidrule{2-13}
     &\multirow{2}{*}{c}&\multirow{2}{*}{EYOC (ours)}	 &- & KITTI &-& \checkmark & \underline{\underline{83.2}}	&99.5 	&96.6 	&89.1 	&\underline{\underline{78.6}} 	&\underline{\underline{52.3}} \\
    &&&WOD & KITTI &-& \checkmark     &80.6 	&99.5 	&95.6 	&89.1 	&75.1 	&43.7  \\  
    % \hline
    % \hline
    % \cmidrule{2-13}
    % \cmidrule{2-13}
    \midrule
    \midrule
    % 
    % 
    % WOD
    % 
    % 
    \multirow{8}{*}{\rotatebox{90}{WOD}}&\multirow{2}{*}{d}&FCGF \cite{choy2019fully}	      &KITTI &-  &\checkmark &-  &69.9 	&97.1 	&87.9 	&61.8 	&59.0 	&43.9  \\
    % &&FCGF + C       &KITTI &-  &\checkmark &\\
    &&Predator \cite{huang2021predator}	  &KITTI &-  &\checkmark & -  &70.7 	&\underline{98.1} 	&\underline{\underline{97.6}} 	&\underline{\underline{81.2}} 	&53.2 	&23.6\\
    % &&GCL \cite{ao2023buffer}           &KITTI &-  &\checkmark & -  &66.9 	&100	&93.7	&68.8	&41.7	&30.1\\
    \cmidrule{2-13}
    &\multirow{3}{*}{e}&FCGF \cite{choy2019fully}      &- & WOD &\checkmark & - &\textbf{89.5} 	&\textbf{100.0} 	&\underline{98.6} 	&\underline{91.2} 	&\textbf{83.5} 	&\textbf{74.0}  \\
    &&FCGF + C       &- & WOD &\checkmark &\checkmark &77.2 	&\underline{98.1} 	&89.9 	&75.8 	&64.7 	&\underline{\underline{57.7}}  \\
    &&Predator \cite{huang2021predator}	   &- & WOD &\checkmark & -  &\underline{86.4} 	&\textbf{100.0} 	&\textbf{100.0} 	&\textbf{95.3} 	&\underline{79.1} 	&57.7 \\
    \cmidrule{2-13}
    &\multirow{2}{*}{f} &\multirow{2}{*}{EYOC (ours)}	  &- & WOD &- &\checkmark &\underline{\underline{78.4}} 	&\underline{\underline{97.6}} 	&91.3 	&78.2 	&\underline{\underline{65.5}} 	&\underline{59.3}  \\
    &&&KITTI & WOD &- & \checkmark     &77.3 	&97.1 	&90.3 	&75.8 	&\underline{\underline{65.5}} 	&\underline{\underline{57.7}}  \\  
    % \hline
    % \hline
    % \cmidrule{2-13}
    % \cmidrule{2-13}
    \midrule
    \midrule
    % 
    % 
    % nuScenes
    % 
    % 
    \multirow{8}{*}{\rotatebox{90}{nuScenes}}&\multirow{2}{*}{g} &FCGF \cite{choy2019fully}   &WOD &-  &\checkmark &-  &\underline{67.1} 	&\underline{98.9} 	&\textbf{93.9} 	&\textbf{73.6} 	&\underline{42.6} 	&\underline{\underline{26.3}}  \\
    % &&FCGF + C	     &WOD &-  &\checkmark & \\
    &&Predator \cite{huang2021predator}	  &WOD &-  &\checkmark & -  &34.5 	&93.0 	&55.2 	&11.8 	&6.0 	&6.7 \\
    \cmidrule{2-13}
    &\multirow{3}{*}{h}&FCGF \cite{choy2019fully}   	      &- & nuScenes & \checkmark & - &39.5 	&87.9 	&63.9 	&23.6 	&11.8 	&10.2 \\
    &&FCGF + C	       &- & nuScenes &\checkmark & \checkmark &59.3 	&96.2 	&85.1 	&59.6 	&35.8 	&20.0 \\
    &&Predator \cite{huang2021predator}  	  &- & nuScenes &\checkmark & -  &51.0 	&\textbf{99.7}	&72.2	&52.8	&16.2	&14.3\\
    % &&GCL-Conv \cite{liu2023density}	&-&nuScenes & - &85.5 	&99.3 	&97.7 	&91.8 	&77.8 	&60.7 \\
    \cmidrule{2-13}
    &\multirow{2}{*}{i}&\multirow{2}{*}{EYOC (ours)} &- & nuScenes &- & \checkmark &\underline{\underline{61.7}} 	&\underline{\underline{96.7}} 	&\underline{\underline{85.6}} &	\underline{61.8} 	&\underline{\underline{37.5}} 	&\underline{26.9} \\
    &                             &&WOD & nuScenes & -& \checkmark     &\textbf{68.4} 	&\underline{98.9} 	&\underline{91.7} 	&\underline{73.3} 	&\textbf{44.3} 	&\textbf{33.7}  \\  
    % \hline
    % \hline
    \bottomrule
  \end{tabular}
  }
%   \vspace{0.1cm}
  \vspace{-0.2cm}
  \caption{\textbf{Comparison of mRR(\%) and RR (\%) between SOTA methods and EYOC over five test sets with $d\in [b_1,b_2]$ on KITTI \cite{Geiger2012CVPR}, WOD \cite{Sun_2020_CVPR}, and nuScenes \cite{Caesar_2020_CVPR}, respectively}, with increasing point cloud distance and registration difficulty. We group the tests denoted by letters \textit{a-i}, \textbf{where \textit{c,f,i} denotes EYOC, \textit{a,d,g} are the fair generalization results of supervised methods and \textit{b,e,h} mark the oracle supervised performance with labels on the new dataset}. EYOC is the only unsupervised method. We use `FCGF + C' to denote FCGF trained with progressive datasets, which is a theoretical upper bound for EYOC. All features are registered using RANSAC.}
  \vspace{-0.3cm}
  \label{tab:comparison}
\end{table*}

% \vspace{-0.1cm}
\paragraph{Training.}
\label{sec:training}
For supervised comparison methods, we follow common practice \cite{ijcai2023p134} to train the model on traditional datasets with $d\in [5,20]$ and further finetune on $d\in [5,50]$. The strategy applies to all baselines, while pretrained weights will be used for those whose training does not converge (denoted with *). On the other hand, EYOC needs only one course of training thanks to the progressive dataset. When a labelled pretraining dataset is available, the parameters of adaptive spatial filtering are acquired with the help of pose labels, presumably from KITTI or WOD; Otherwise, we use hard spatial filtering. The complete training of EYOC consists of 200 epochs with 0.001 learning rate and $1\times 10^{-4}$ weight decay, same as FCGF, implemented with MinkowskiEngine \cite{choy20194d} and Pytorch3D \cite{ravi2020pytorch3d}.

\paragraph{Inference.} When conducting a comparison with previous methods, we apply RANSAC \cite{fischler1981random} on all methods including EYOC for fairness. Otherwise, we default EYOC inference to SC$^2$-PCR \cite{chen2022sc2} for speed and performance.
\vspace{-0.2cm}

\paragraph{Metrics.} We report 5 metrics according to existing literature \cite{gojcic2018learned,choy2019fully,ijcai2023p134}: Registration Recall (RR), Relative Rotation Error (RRE), Relative Translation Error (RTE), Mean RR (mRR), and Inlier Ratio (IR), the formal definition of which can be found in Appendix \cref{sec:appendix_metric}. We apply IR \textit{on the generated labeler correspondences} to indicate their quality during training.
% \vspace{-0.2cm}

% \paragraph{Supervised generalization versus unsupervised finetuning.} Given a pretrained model and a new out-of-distribution dataset, unsupervised methods can perform finetuning while supervised methods cannot, causing a performance increase. In that case, the similarity records used by the adaptive spatial filtering is collected on the pretraining dataset to avoid label leakage.

\subsection{Overall Performance}
We compare both a generalization setting (\textit{a,d,g}) and finetuning setting (\textit{b,e,h}) for SOTA supervised methods, against the unsupervised EYOC (\textit{c,f,i}) on three datasets, KITTI \cite{Geiger2012CVPR}, WOD \cite{Sun_2020_CVPR}, and nuScenes \cite{Caesar_2020_CVPR}, respectively in \cref{tab:comparison}.

\begin{table*}[t]
  \centering
  \small
%   \hspace{-0.3cm}
\begin{minipage}{0.56\linewidth}
  \resizebox{\linewidth}{!}{
  \begin{tabular}{@{}lccccc|c|cccc@{}}
    \toprule
    &&&&&& 1st Epoch &&\multicolumn{3}{c}{$[40,50]$} \\
    \cline{9-11}
    No.&LF & SF-h & SF-a & SR+CR & PD & Labeler IR & mRR & RR & RRE & RTE \\
    \midrule
    a & - & - & - & - & \checkmark                   & 5.1 &  \multicolumn{4}{c}{\multirow{6}{*}{N/C}} \\
    b & \checkmark & - & - & - & \checkmark          & 1.5 &  &  &  &\\
    c & \checkmark & - & - & \checkmark & \checkmark & 0.6 &  &  &  & \\
    d & \checkmark & - & \checkmark & - & \checkmark & 5.9 &  &  &  & \\
    e & \checkmark & \checkmark & - & - & \checkmark & 5.9 &  &  &  & \\
    f & - & - & \checkmark & \checkmark & -          & 0.0 &  &  &  & \\
    \midrule
    % \midrule
    % &&&&&&& 1st Epoch &\multicolumn{3}{c}{$[40,50]$} \\
    % \cmidrule{9-11}
    % No.&LF & SF (hard) & SF (adaptive) & SR & PD & mRR  & Labeler IR & RR & RRE & RTE \\
    g & \checkmark & - & \checkmark & \checkmark & \checkmark &7.8 	&87.5	            &66.8	            &\textbf{1.3}	&\textbf{29.7}\\
    h & - & - & - & \checkmark & \checkmark                   &18.4	&84.6	            &60.3	            &1.4	&33.9\\
    i & - & - & \checkmark & \checkmark & \checkmark          &\underline{43.3} & \textbf{88.0} 	&\textbf{68.8}	    &\textbf{1.3}	&\underline{31.8}\\
    j & - & \checkmark & - & \checkmark & \checkmark          &\textbf{53.2} & \underline{87.6} 	&\underline{67.8}	&\underline{1.31}	&32.2 \\
    \bottomrule
  \end{tabular}
  }
\end{minipage}
\hspace{-0.1cm}
\quad
\begin{minipage}{0.095\linewidth}
    \resizebox{1\linewidth}{!}{
    \begin{tabular}{@{}lc@{}}
    \toprule
    $\lambda$ & [40,50]\\
    \midrule
    0.0 &  71.9 \\
    0.1 &  70.4 \\
    0.2 &  \textbf{73.9} \\
    0.3 &  71.4 \\
    0.4 &  69.3 \\
    0.5 &  71.9 \\
    0.6 &  69.8 \\
    0.7 &  \underline{72.9} \\
    0.8 &  67.3 \\
    0.85 & 58.8  \\
    0.9 &  N/C  \\
    0.99 & N/C   \\
    \bottomrule
  \end{tabular}
  }
\end{minipage}
% \hfill
\hspace{-0.3cm}
\quad
\begin{minipage}{0.145\linewidth}
    \resizebox{1\linewidth}{!}{
    \begin{tabular}{@{}lc@{}}
    \toprule
    & 1st Epoch\\
    $d_{thresh}$ & Labeler IR\\
    \midrule
    0  & 18.4  \\
    5  & 18.5  \\
    10 & 25.2  \\
    15 & 31.1  \\
    20 & 31.4  \\
    25 & 29.4  \\
    30 & 45.1  \\
    35 & \underline{49.0}  \\
    40 & \textbf{53.2}  \\
    45 & 43.3  \\
    \bottomrule
  \end{tabular}
  }
\end{minipage}
\hspace{-0.3cm}
\quad
\begin{minipage}{0.145\linewidth}
    \resizebox{1\linewidth}{!}{
    \begin{tabular}{@{}lc@{}}
    \toprule
    & 1st Epoch\\
    $s_{thresh}$ & Labeler IR\\
    \midrule
    0.0 & 18.4  \\
    0.1 & 25.4  \\
    0.2 & 30.7  \\
    0.3 & 31.5  \\
    0.4 & 31.1  \\
    0.5 & \underline{34.9}  \\
    0.6 & \textbf{43.3}  \\
    0.7 & N/C  \\
    0.8 & N/C  \\
    0.9 & N/C  \\
    \bottomrule
  \end{tabular}
  }
\end{minipage}

%   }
%   \vspace{0.1cm}
  \vspace{-0.2cm}
  \caption{\textbf{Ablation study of EYOC.} Labeler IR (\%), mRR (\%), RR@$[40,50]$ (\%), RRE (\degree), and RTE (cm) on KITTI val set are presented. Lowe Filtering (LF), Spatial Filtering of \textit{hard} (SF-h) or \textit{adaptive} (SF-a) strategies, Speculative Registration and Correspondence Rediscovery (SR+CR), progressive Dataset (PD), EMA decay factor $\lambda$, and two parameters of Spatial Filtering, $d_{thresh},s_{thresh}$, are ablated.}
  \label{tab:ablation}
  \vspace{-0.5cm}
\end{table*}

We first notice that supervised methods do fail to generalize to different datasets, according to \textit{a-b,d-e,} and \textit{g-h} in \cref{tab:comparison}. 
% The bigger the domain gap between training and testing datasets, the worse supervised methods perform. 
Generalizing from WOD to KITTI, which are both 64-line datasets with small domain shift, supervised methods suffer $5.6\%$ and $15.6\%$ mRR drop respectively for FCGF \cite{choy2019fully} and Predator \cite{huang2021predator}, when compared with models trained on KITTI directly (rows \textit{a} and \textit{b}). Similar results are seen generalizing from KITTI to WOD as well (rows \textit{d} and \textit{e}), with $19.6\%$ and $15.7\%$ mRR drop for FCGF and Predator, respectively. On the other hand, a harder dataset, nuScenes with only a 32-laser LiDAR, struggles to support supervised training. We witness worse supervised performance than generalization scores from WOD for FCGF when comparing the rows \textit{g} and \textit{h}. Nonetheless, their generalization scores are also subpar, hitting merely $67.1\%$ and $34.5\%$ mRR with FCGF and Predator, respectively on nuScenes in row \textit{g}. Additionally, contrary to the common belief, Predator performs much worse than FCGF on an out-of-domain dataset, nuScenes, in row \textit{g}.

\paragraph{Does unsupervised finetuning improve upon supervised methods on out-of-domain unlabelled data?} By comparing \textit{a-c}, \textit{d-f}, and \textit{g-i} in \cref{tab:comparison}, we confirm that EYOC improves upon fixed supervised models by a considerable margin through unsupervised finetuning. On KITTI, EYOC surpasses raw FCGF, achieving $83.2\%(+11.4\%)$ and $80.6\%(+8.8\%)$ mRR by training from scratch and finetuning, respectively. On WOD and nuScenes, the respective figures are $78.4\%(+8.5\%)$ and $77.3\%(+7.4\%)$ on WOD, $61.7\%(-5.3\%)$ and $68.4\%(+1.3\%)$ on nuScenes compared to FCGF. We conclude that, given a pretrained model and an incoming unlabelled dataset, applying EYOC for unsupervised training/finetuning provides a considerable performance boost on the new dataset.

\paragraph{Is EYOC comparable to supervised methods on labelled data?} Unsupervised methods have to perform similarly to supervised ones in order to be considered valuable. Through comparing \textit{b-c}, \textit{e-f}, and \textit{h-i} in \cref{tab:comparison}, we find that EYOC exhibits comparable performance with SOTA fully-supervised methods when trained on the same dataset. On KITTI, mRR of EYOC is only $4.7\%$ and $1.4\%$ lower than that of the best-performing Predator and FCGF+C, respectively. Other low-overlap registration methods, excluding CoFiNet \cite{yu2021cofinet}, are less suitable for outdoor scenarios, as SpinNet \cite{ao2021spinnet}, D3Feat \cite{bai2020d3feat}, and Geotransformer \cite{qin2022geometric} suffer from divergence. In the meantime, different results are reported on WOD where EYOC is $10.9\%$ behind FCGF but $1.2\%$ ahead of FCGF+C, indicating that FCGF+C is not always effective on all datasets. EYOC echibits stronger results on nuScenes, surpassing FCGF+C by $9.9\%$ instead. We conclude that EYOC does perform similarly to fully supervised methods while requiring no pose labels at all.

% \paragraph{Does our progressive dataset help baselines as well?} 

\subsection{Ablation}
\label{sec:ablation}
\paragraph{Structural components.} We first ablate supporting structures of EYOC in \cref{tab:ablation}, including Lowe Filtering (LF), Spatial Filtering with both \textit{hard} (SF-h) and \textit{adaptive} (SF-a) strategies, Speculative Registration + Correspondence Rediscovery (SR+CR), and the Progressive Dataset (PD). Judging from \textit{a-b} and \textit{g-i}, Lowe's filter deteriorates IR by $3.60\%$ and $35.5\%$, respectively, contrary to previous findings on indoor RGB-D images. We keep Lowe filtering as an option in case of other datasets. Also, lone Spatial Filtering or speculative registration both fail to support training according to \textit{c,d,e}. The best-performing setup (\textit{i}) fails completely without Progressive Dataset (\textit{f}) at $0.0\%$ IR, indicating the importance of the Progressive Dataset strategy. On the other hand, converged setups reveal consistently higher IR up to $53.2\%$. SF-h (\textit{i}) and SF-s (\textit{j}) achieve $88.0\%$ and $87.6\%$ mRR, respectively, slightly better than not using Spatial Filtering (\textit{h}) which achieves $84.6\%$ mRR. Similar trends are observed with respective performance on long-range pairs as well, where \textit{i} and \textit{j} outperforms \textit{h} by $8.5\%$ and $7.5\%$ RR, respectively. We default EYOC structure to SF-a, SR+CR, and PD (\textit{i}) for the highest performance.

\paragraph{Parameter choices.} Three parameter choices, $\lambda$, $d_{thresh}$, and $s_{thresh}$, are discussed in \cref{tab:ablation} as well. For the EMA decay factor $\lambda$, any value less than $0.7$ achieves similar results averaging at $71.4\%$, while larger $\lambda$ quickly drains the performance. On the other hand, similar to our previous findings \cref{sec:hypothesis}, IR marks better scores with stricter thresholds of $d_{thresh}$ and $s_{thresh}$ (\textit{i.e.}, using regions farther from the LiDAR), but the number of correspondences could shrink to the point of causing divergence under an extreme threshold. In light of this phenomenon, we choose $\lambda=0.2$, $d_{thresh}=40m$, and $s_{thresh}=0.6$ are default parameters for the best performance just before the divergence line. Should a divergence occur on new datasets, these thresholds could be lowered to cater to new data distributions.

\begin{figure}[t]
  \centering
%   \vspace{-0.2cm}
  \includegraphics[width=0.8\linewidth]{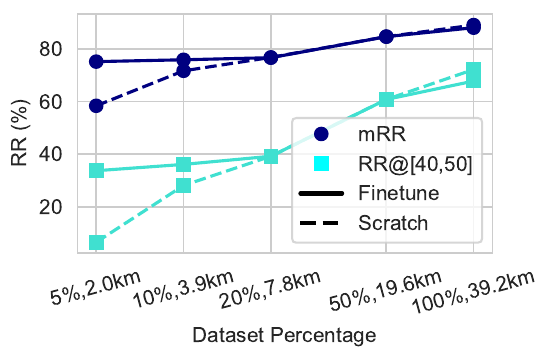}
  \vspace{-0.3cm}
  \caption{\textbf{Comparison between finetuning from WOD and training from scratch for EYOC}, with the first 5\% to 100\% of unlabelled KITTI, where both RR on $d\in [40,50]$ and mRR are displayed. The horizontal axis is in log scale. Finetuning exhibits more stability before 20\%, while training from scratch performs better after 50\%.}
  \label{fig:percentage}
%   \vspace{-0.3cm}
\end{figure}

\begin{table}[]
    \centering
    % \resizebox{\linewidth}{!}{
    %   \begin{tabular}{@{}lccccc|c@{}}
    %     \toprule
    %       & \multicolumn{5}{c|}{Training (one pass)} & \# Training\\
    %       \cmidrule{2-6}
    %        & Data &Feature & Label Gen. & Loss & Total& Required\\
    %     %   & Loading &Extracton & Generation & Prop. & Passes\\
    %     \midrule
    %     FCGF \cite{choy2019fully}   &692 &128 & - &356 &1176 &2\\
    %     EYOC                        &18 &170 &381 &296 &865 &1\\
    %     \bottomrule
    %   \end{tabular}
    %  }
    \resizebox{\linewidth}{!}{
    \begin{tabular}{@{}lcccccc|c@{}}
      \toprule
        & \multicolumn{6}{c|}{Training (one pass)} & \# Training\\
        \cmidrule{2-7}
         & Data & NN-S & Feat. & Label Gen. & Loss & Total& Required\\
      %   & Loading &Extracton & Generation & Prop. & Passes\\
      \midrule
      FCGF \cite{choy2019fully}   &692 & - &128 & - &356 &1176 &$\times$2\\
      FCGF*   &17 & 33 &152 & - &301 &503 &$\times$2\\
      EYOC    &18 & - &170 &381 &296 &865 &$\times$1\\
      \bottomrule
    \end{tabular}
   }
     % \vspace{-0.2cm}
    \caption{Time analysis of EYOC, FCGF \cite{choy2019fully}, and FCGF with GPU-accelerated NN-Search (denoted with *) in milliseconds. The number of complete training routines required for a network is listed in the last column.}
    \label{tab:time}
    % \vspace{-0.5cm}
\end{table}

\subsection{Finetuning versus From Scratch}
We further compare the finetuning and training-from-scratch strategies for EYOC with different portions of the new dataset KITTI, while assuming a pretrained model on WOD is available. Metrics including RR @ $d\in [40,50]$, mRR, and driving distance (km) on KITTI are displayed with the first $5\%$ to $100\%$ of KITTI, as illustrated in \cref{fig:percentage}. Overall, performance of both methods increase with the amount of training data; However, finetuning grants more stability by inheriting knowledge from the previous dataset, therefore performing better with smaller datasets then $20\%$ ($7.8km$) where the mRR stablizes around $75\%$. On the other hand, training from scratch achieves better results after $50\%$ ($19.6km$), peaking at the full dataset with $89.1\%$ mRR and $72.2\%$ RR @ $d\in [40,50]$, respectively. We conclude that finetuning is better for datasets roughly shorter than 10km, while training from scratch would be a better choice for larger datasets.

\subsection{Time Analysis}
\label{sec:time}
We break down the training time for FCGF \cite{choy2019fully} and EYOC in \cref{tab:time}. Because the NN-search in Correspondence Rediscovery of EYOC is accelerated with GPU using Pytorch3D \cite{ravi2020pytorch3d}, it is necessary to apply the same trick to the baseline FCGF for fair comparison, which is termed `FCGF*'. While EYOC needs an additional $381$ms for label generation, it completes training once and for all, resulting in the lowest total training time. On the other hand, FCGF* is trained twice to prevent divergence \cite{ijcai2023p134} as detailed in \cref{sec:training}. In comparison, vanilla FCGF ranks the slowest due to a prolonged data loading time of 692ms. We conclude that EYOC enjoys a lower training cost than its supervised counterpart.

\begin{figure}[t]
  \centering
%   \vspace{-0.2cm}
  \includegraphics[width=0.45\linewidth]{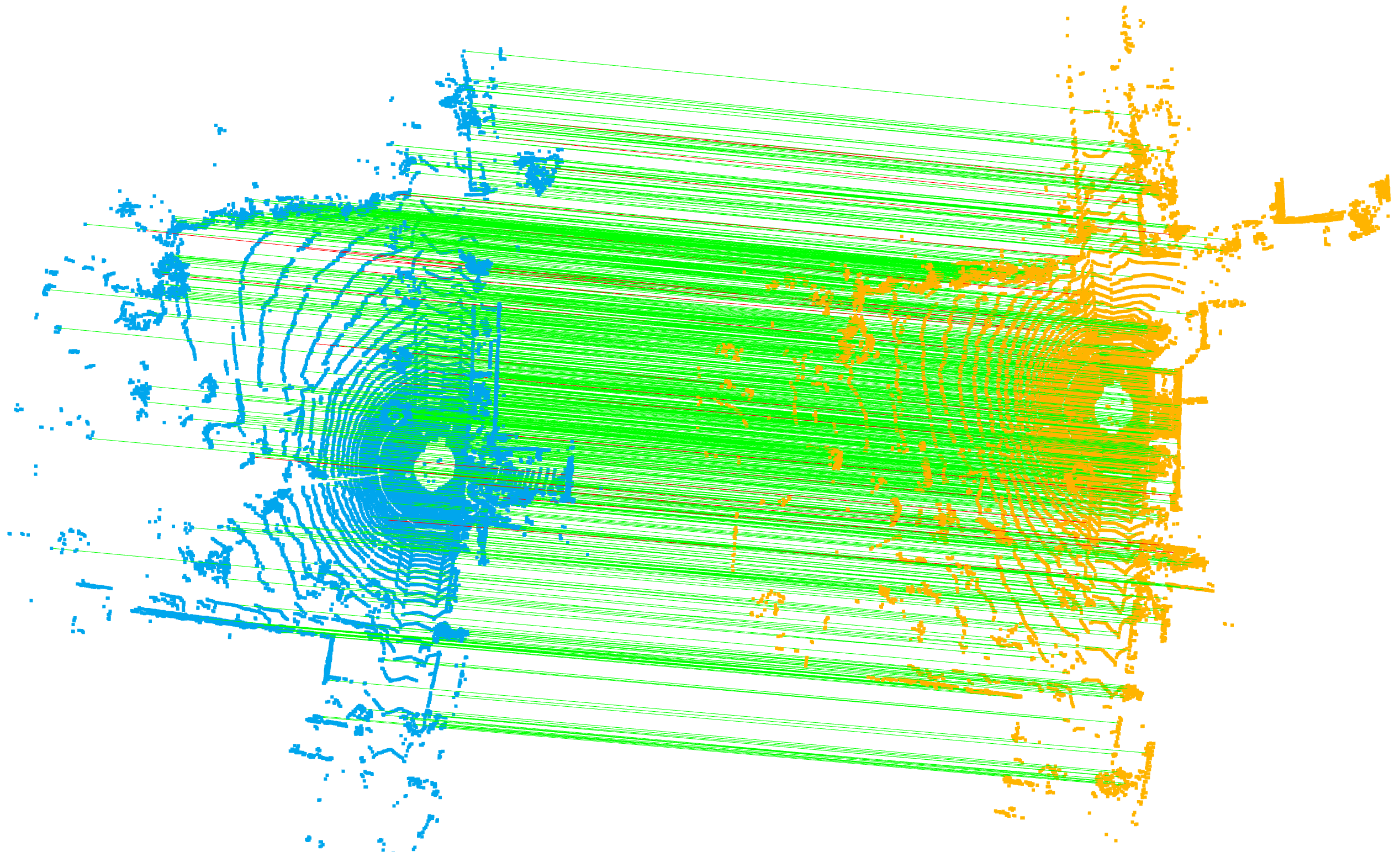}
  \quad
  \includegraphics[width=0.3\linewidth]{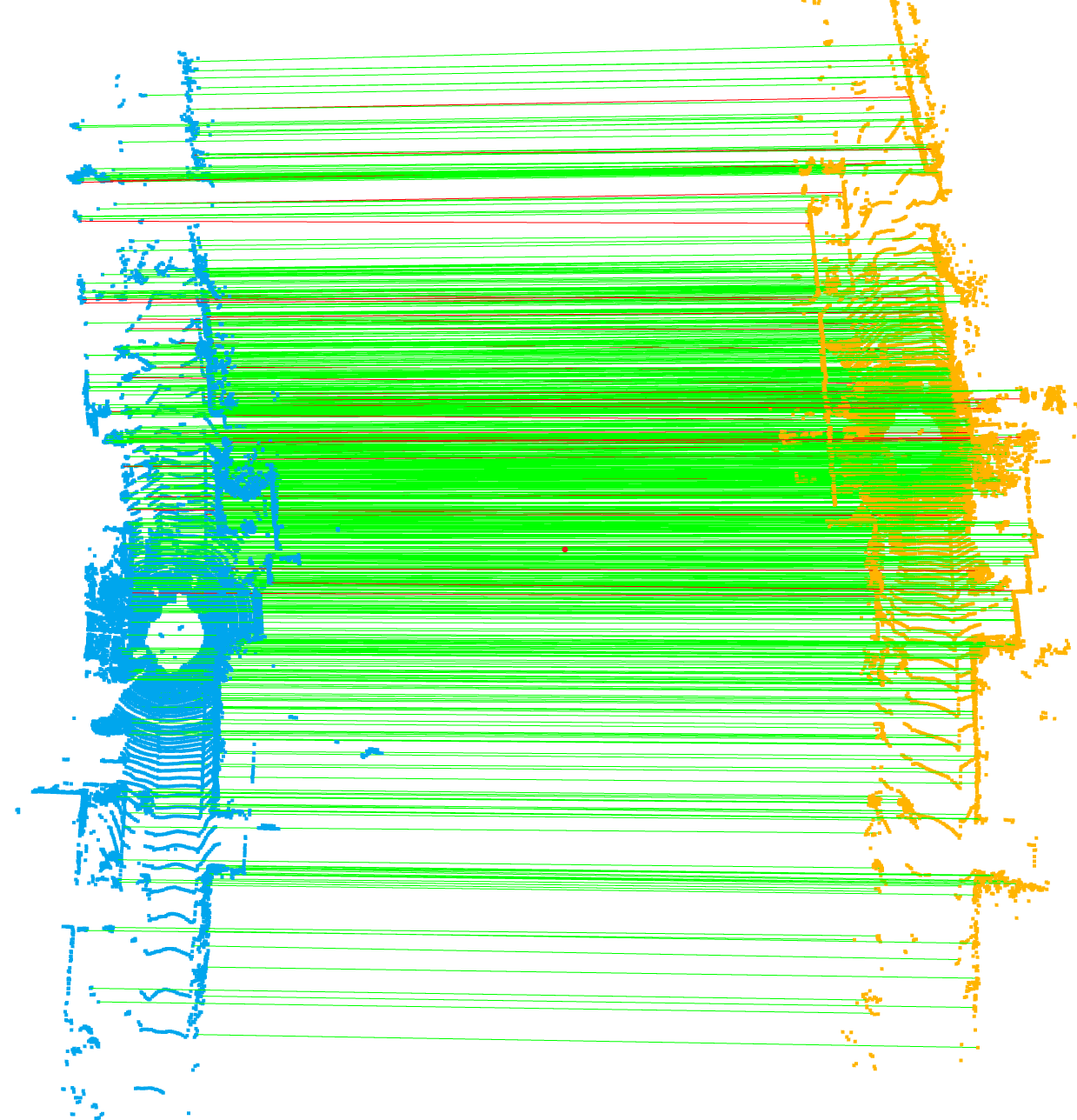}
\vspace{-0.3cm}

  \includegraphics[width=0.45\linewidth]{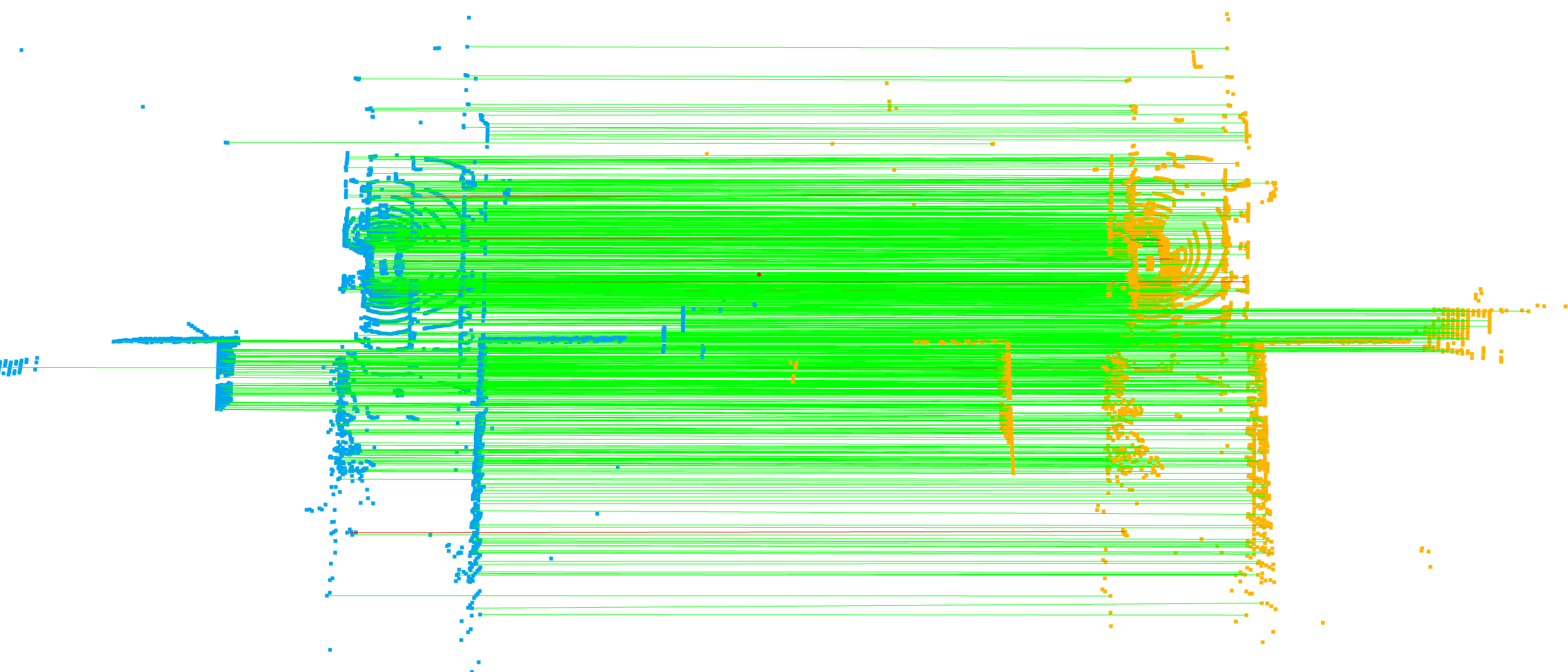}
  \quad
  \includegraphics[width=0.45\linewidth]{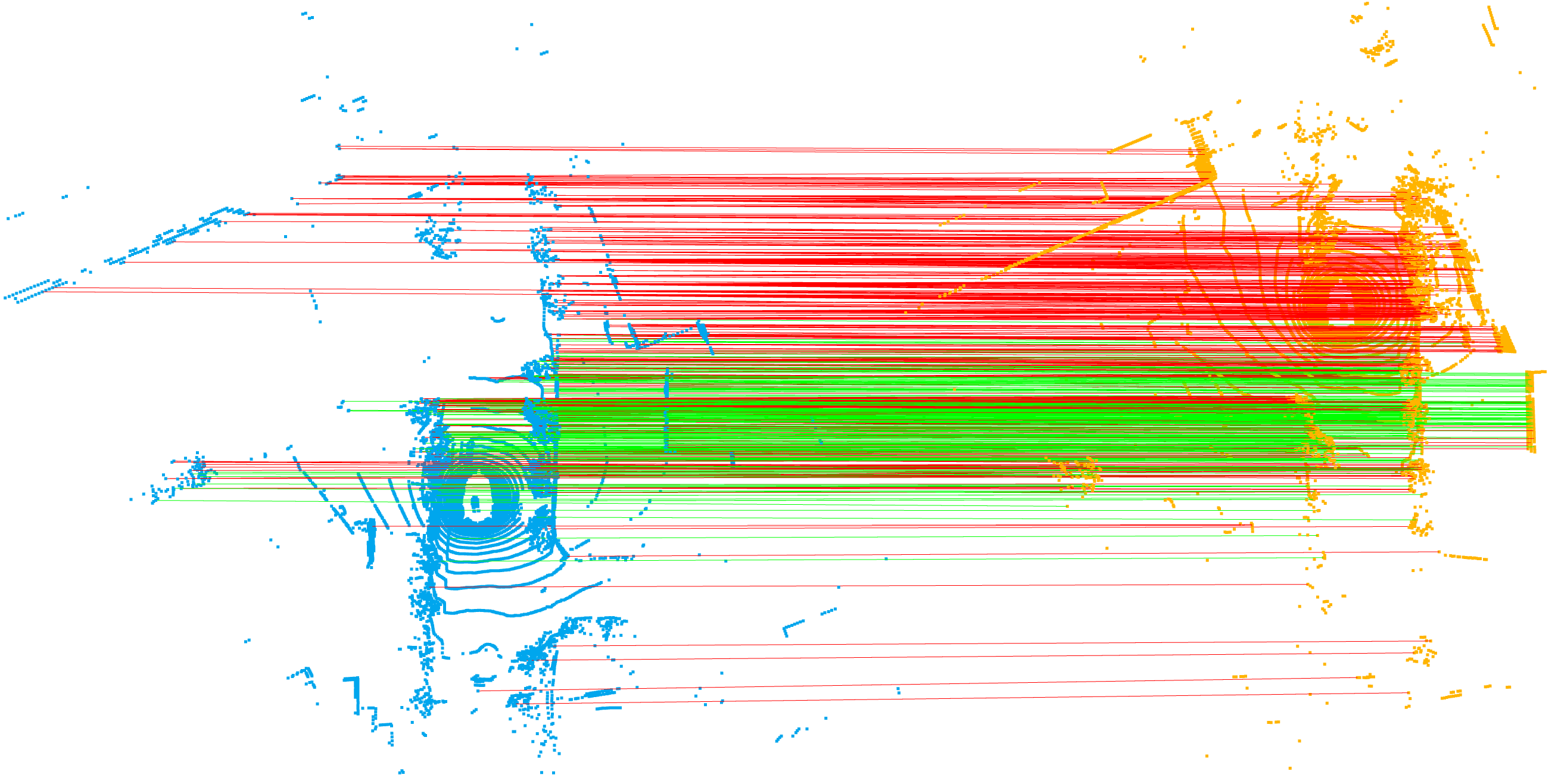}
\vspace{-0.3cm}

  \includegraphics[width=0.45\linewidth]{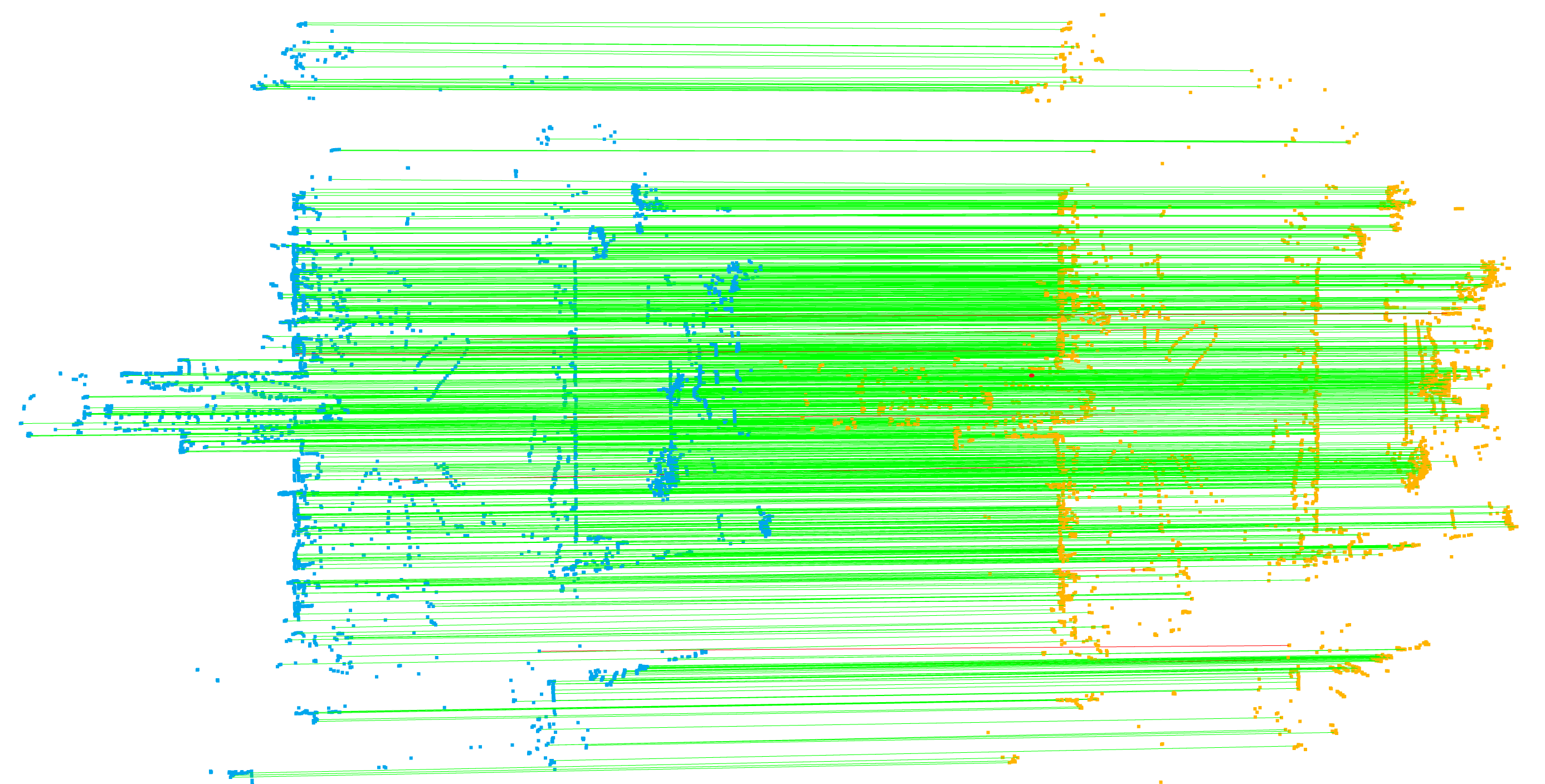}
  \quad
  \includegraphics[width=0.45\linewidth]{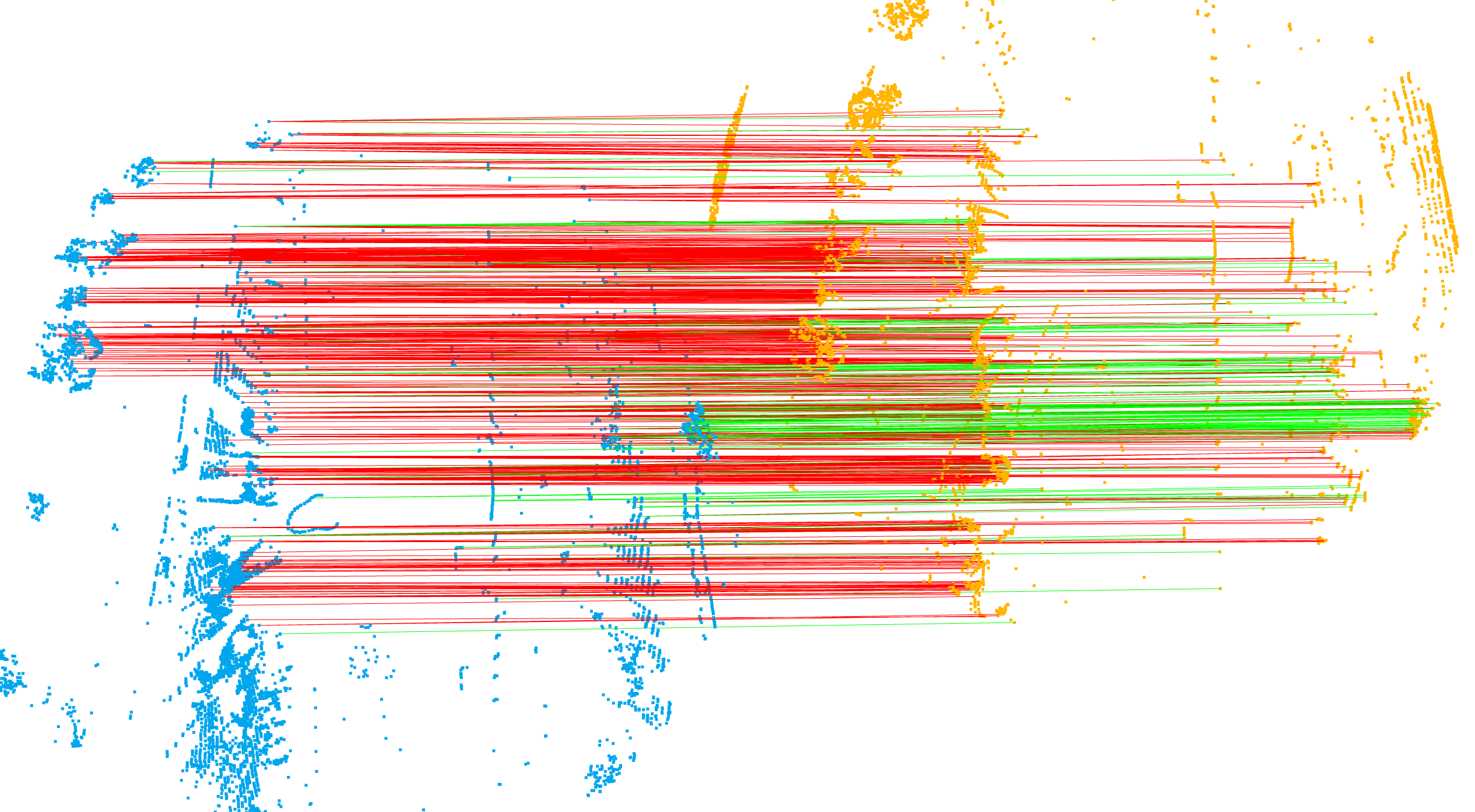}

  \vspace{-0.3cm}

  \caption{\textbf{Visualization of clean correspondence labels on KITTI (top row), nuScenes (middle row), and WOD (bottom row)}, where correspondences with $\leq1m$ location error are colored green and otherwise red. Even when Speculative Registration fails, most of the false correspondences are in parallel to correct ones, they are just not precise but still informative.}
  \label{fig:visualization}
  \vspace{-0.3cm}
\end{figure}

% \subsection{Visualization}

\section{Conclusion}

We have proposed EYOC, an unsupervised distant point cloud registration technique that requires nothing more than consecutive LiDAR sweeps, which is easily acquired on-the-fly with self-driving vehicles. With the correspondence filtering pipeline built upon our investigations, EYOC allows a 3D feature extractor to generate labels for itself, enabling fully unsupervised training. Extensive experiments demonstrate that, while enjoying comparable performance to supervised methods, EYOC also has a lower training cost, thus being preferable compared to the traditional `manual labelling + supervised training' paradigm. EYOC's unrivalled capability of finetuning on new data distributions marks a step towards the mass deployment of collaborative sensing on SDVs.

\paragraph{Acknowledgement.} This work was supported in part by the Natural Science Foundation of Shanghai (Grant No.22ZR1400200) and the Fundamental Research Funds for the Central Universities (No. 2232023Y-01).

% \clearpage
{
    \small
    \bibliographystyle{ieeenat_fullname}
    \bibliography{main}

\begin{thebibliography}{61}
\providecommand{\natexlab}[1]{#1}
\providecommand{\url}[1]{\texttt{#1}}
\expandafter\ifx\csname urlstyle\endcsname\relax
  \providecommand{\doi}[1]{doi: #1}\else
  \providecommand{\doi}{doi: \begingroup \urlstyle{rm}\Url}\fi

\bibitem[Ali et~al.(2021)Ali, Kahraman, Reis, and Stricker]{ali2021rpsrnet}
Sk~Aziz Ali, Kerem Kahraman, Gerd Reis, and Didier Stricker.
\newblock {RPSRNet}: End-to-end trainable rigid point set registration network
  using {Barnes-Hut 2D-Tree} representation.
\newblock In \emph{Proceedings of the IEEE/CVF Conference on Computer Vision
  and Pattern Recognition}, pages 13100--13110, 2021.

\bibitem[Ao et~al.(2021)Ao, Hu, Yang, Markham, and Guo]{ao2021spinnet}
Sheng Ao, Qingyong Hu, Bo Yang, Andrew Markham, and Yulan Guo.
\newblock {SpinNet}: Learning a general surface descriptor for {3D} point cloud
  registration.
\newblock In \emph{Proceedings of the IEEE/CVF Conference on Computer Vision
  and Pattern Recognition}, pages 11753--11762, 2021.

\bibitem[Ao et~al.(2023)Ao, Hu, Wang, Xu, and Guo]{ao2023buffer}
Sheng Ao, Qingyong Hu, Hanyun Wang, Kai Xu, and Yulan Guo.
\newblock {BUFFER}: Balancing accuracy, efficiency, and generalizability in
  point cloud registration.
\newblock In \emph{Proceedings of the IEEE/CVF Conference on Computer Vision
  and Pattern Recognition}, pages 1255--1264, 2023.

\bibitem[Bai et~al.(2020)Bai, Luo, Zhou, Fu, Quan, and Tai]{bai2020d3feat}
Xuyang Bai, Zixin Luo, Lei Zhou, Hongbo Fu, Long Quan, and Chiew-Lan Tai.
\newblock {D3Feat}: Joint learning of dense detection and description of {3D}
  local features.
\newblock In \emph{Proceedings of the IEEE/CVF Conference on Computer Vision
  and Pattern Recognition}, pages 6359--6367, 2020.

\bibitem[Bai et~al.(2021)Bai, Luo, Zhou, Chen, Li, Hu, Fu, and
  Tai]{bai2021pointdsc}
Xuyang Bai, Zixin Luo, Lei Zhou, Hongkai Chen, Lei Li, Zeyu Hu, Hongbo Fu, and
  Chiew-Lan Tai.
\newblock {PointDSC}: Robust point cloud registration using deep spatial
  consistency.
\newblock In \emph{Proceedings of the IEEE/CVF Conference on Computer Vision
  and Pattern Recognition}, pages 15859--15869, 2021.

\bibitem[Caesar et~al.(2020)Caesar, Bankiti, Lang, Vora, Liong, Xu, Krishnan,
  Pan, Baldan, and Beijbom]{Caesar_2020_CVPR}
Holger Caesar, Varun Bankiti, Alex~H. Lang, Sourabh Vora, Venice~Erin Liong,
  Qiang Xu, Anush Krishnan, Yu Pan, Giancarlo Baldan, and Oscar Beijbom.
\newblock {nuScenes}: A multimodal dataset for autonomous driving.
\newblock In \emph{Proceedings of the IEEE/CVF Conference on Computer Vision
  and Pattern Recognition}, 2020.

\bibitem[Chen et~al.(2022)Chen, Sun, Yang, and Tao]{chen2022sc2}
Zhi Chen, Kun Sun, Fan Yang, and Wenbing Tao.
\newblock {SC2-PCR}: A second order spatial compatibility for efficient and
  robust point cloud registration.
\newblock In \emph{Proceedings of the IEEE/CVF Conference on Computer Vision
  and Pattern Recognition}, pages 13221--13231, 2022.

\bibitem[Choy et~al.(2019{\natexlab{a}})Choy, Gwak, and Savarese]{choy20194d}
Christopher Choy, JunYoung Gwak, and Silvio Savarese.
\newblock 4d spatio-temporal convnets: Minkowski convolutional neural networks.
\newblock In \emph{Proceedings of the IEEE/CVF Conference on Computer Vision
  and Pattern Recognition}, pages 3075--3084, 2019{\natexlab{a}}.

\bibitem[Choy et~al.(2019{\natexlab{b}})Choy, Park, and Koltun]{choy2019fully}
Christopher Choy, Jaesik Park, and Vladlen Koltun.
\newblock Fully convolutional geometric features.
\newblock In \emph{Proceedings of the IEEE/CVF International Conference on
  Computer Vision}, 2019{\natexlab{b}}.

\bibitem[Choy et~al.(2020)Choy, Dong, and Koltun]{choy2020deep}
Christopher Choy, Wei Dong, and Vladlen Koltun.
\newblock Deep global registration.
\newblock In \emph{Proceedings of the IEEE/CVF Conference on Computer Vision
  and Pattern Recognition}, pages 2514--2523, 2020.

\bibitem[Deng et~al.(2018)Deng, Birdal, and Ilic]{deng2018ppfnet}
Haowen Deng, Tolga Birdal, and Slobodan Ilic.
\newblock {PPFNet}: Global context aware local features for robust {3D} point
  matching.
\newblock In \emph{Proceedings of the IEEE/CVF Conference on Computer Vision
  and Pattern Recognition}, pages 195--205, 2018.

\bibitem[El~Banani and Johnson(2021)]{el2021bootstrap}
Mohamed El~Banani and Justin Johnson.
\newblock Bootstrap your own correspondences.
\newblock In \emph{Proceedings of the IEEE/CVF International Conference on
  Computer Vision}, pages 6433--6442, 2021.

\bibitem[El~Banani et~al.(2021)El~Banani, Gao, and
  Johnson]{el2021unsupervisedr}
Mohamed El~Banani, Luya Gao, and Justin Johnson.
\newblock {UnsupervisedR\&R}: Unsupervised point cloud registration via
  differentiable rendering.
\newblock In \emph{Proceedings of the IEEE/CVF Conference on Computer Vision
  and Pattern Recognition}, pages 7129--7139, 2021.

\bibitem[Fischler and Bolles(1981)]{fischler1981random}
Martin~A Fischler and Robert~C Bolles.
\newblock Random sample consensus: a paradigm for model fitting with
  applications to image analysis and automated cartography.
\newblock \emph{Communications of the ACM}, 24\penalty0 (6):\penalty0 381--395,
  1981.

\bibitem[Fu et~al.(2021)Fu, Liu, Luo, and Wang]{fu2021robust}
Kexue Fu, Shaolei Liu, Xiaoyuan Luo, and Manning Wang.
\newblock Robust point cloud registration framework based on deep graph
  matching.
\newblock In \emph{Proceedings of the IEEE/CVF conference on computer vision
  and pattern recognition}, pages 8893--8902, 2021.

\bibitem[Geiger et~al.(2012)Geiger, Lenz, and Urtasun]{Geiger2012CVPR}
Andreas Geiger, Philip Lenz, and Raquel Urtasun.
\newblock Are we ready for autonomous driving? the {KITTI} vision benchmark
  suite.
\newblock In \emph{Proceedings of the IEEE/CVF Conference on Computer Vision
  and Pattern Recognition}, 2012.

\bibitem[Giraud et~al.(2017)Giraud, Ta, Bugeau, Coup{\'e}, and
  Papadakis]{giraud2017superpatchmatch}
R{\'e}mi Giraud, Vinh-Thong Ta, Aur{\'e}lie Bugeau, Pierrick Coup{\'e}, and
  Nicolas Papadakis.
\newblock {SuperPatchMatch}: An algorithm for robust correspondences using
  superpixel patches.
\newblock \emph{IEEE Transactions on Image Processing}, 26\penalty0
  (8):\penalty0 4068--4078, 2017.

\bibitem[Gojcic et~al.(2018)Gojcic, Zhou, and Wieser]{gojcic2018learned}
Zan Gojcic, Caifa Zhou, and Andreas Wieser.
\newblock Learned compact local feature descriptor for tls-based geodetic
  monitoring of natural outdoor scenes.
\newblock \emph{International Archives of the Photogrammetry, Remote Sensing
  and Spatial Information Sciences}, 4:\penalty0 113--120, 2018.

\bibitem[Gojcic et~al.(2019)Gojcic, Zhou, Wegner, and
  Wieser]{gojcic2019perfect}
Zan Gojcic, Caifa Zhou, Jan~D Wegner, and Andreas Wieser.
\newblock The perfect match: {3D} point cloud matching with smoothed densities.
\newblock In \emph{Proceedings of the IEEE/CVF Conference on Computer Vision
  and Pattern Recognition}, pages 5545--5554, 2019.

\bibitem[Huang et~al.(2021)Huang, Gojcic, Usvyatsov, Wieser, and
  Schindler]{huang2021predator}
Shengyu Huang, Zan Gojcic, Mikhail Usvyatsov, Andreas Wieser, and Konrad
  Schindler.
\newblock {PREDATOR}: Registration of {3D} point clouds with low overlap.
\newblock In \emph{Proceedings of the IEEE/CVF Conference on Computer Vision
  and Pattern Recognition}, pages 4267--4276, 2021.

\bibitem[Jens et~al.(2019)Jens, Martin, Andres, Jan, Sven, Cyrill, and
  Jurgen]{behley2019iccv}
Behley Jens, Garbade Martin, Milioto Andres, Quenzel Jan, Behnke Sven,
  Stachniss Cyrill, and Gall Jurgen.
\newblock {SemanticKITTI}: A dataset for semantic scene understanding of
  {LiDAR} sequences.
\newblock In \emph{Proceedings of the IEEE/CVF International Conference on
  Computer Vision}, 2019.

\bibitem[Johnson and Hebert(1999)]{johnson1999using}
Andrew~E Johnson and Martial Hebert.
\newblock Using spin images for efficient object recognition in cluttered {3D}
  scenes.
\newblock \emph{IEEE Transactions on Pattern Analysis and Machine
  Intelligence}, 21\penalty0 (5):\penalty0 433--449, 1999.

\bibitem[Lawin et~al.(2018)Lawin, Danelljan, Khan, Forss{\'e}n, and
  Felsberg]{lawin2018density}
Felix~J{\"a}remo Lawin, Martin Danelljan, Fahad~Shahbaz Khan, Per-Erik
  Forss{\'e}n, and Michael Felsberg.
\newblock Density adaptive point set registration.
\newblock In \emph{Proceedings of the IEEE/CVF Conference on Computer Vision
  and Pattern Recognition}, pages 3829--3837, 2018.

\bibitem[Lee et~al.(2021{\natexlab{a}})Lee, Hamsici, Feng, Sharma, and
  Gernoth]{lee2021deeppro}
Donghoon Lee, Onur~C Hamsici, Steven Feng, Prachee Sharma, and Thorsten
  Gernoth.
\newblock {DeepPRO}: Deep partial point cloud registration of objects.
\newblock In \emph{Proceedings of the IEEE/CVF International Conference on
  Computer Vision}, pages 5683--5692, 2021{\natexlab{a}}.

\bibitem[Lee et~al.(2021{\natexlab{b}})Lee, Kim, Cho, and Park]{lee2021deep}
Junha Lee, Seungwook Kim, Minsu Cho, and Jaesik Park.
\newblock Deep hough voting for robust global registration.
\newblock In \emph{Proceedings of the IEEE/CVF International Conference on
  Computer Vision}, pages 15994--16003, 2021{\natexlab{b}}.

\bibitem[Li and Harada(2022)]{li2022lepard}
Yang Li and Tatsuya Harada.
\newblock Lepard: Learning partial point cloud matching in rigid and deformable
  scenes.
\newblock In \emph{Proceedings of the IEEE/CVF Conference on Computer Vision
  and Pattern Recognition}, pages 5554--5564, 2022.

\bibitem[Li et~al.(2022)Li, Wang, Li, Xie, Sima, Lu, Qiao, and
  Dai]{li2022bevformer}
Zhiqi Li, Wenhai Wang, Hongyang Li, Enze Xie, Chonghao Sima, Tong Lu, Yu Qiao,
  and Jifeng Dai.
\newblock {BEVFormer}: Learning bird’s-eye-view representation from
  multi-camera images via spatiotemporal transformers.
\newblock In \emph{European Conference on Computer Vision}, pages 1--18.
  Springer, 2022.

\bibitem[Liu et~al.(2023{\natexlab{a}})Liu, Zhou, Zhu, Chang, and
  Guo]{ijcai2023p134}
Quan Liu, Yunsong Zhou, Hongzi Zhu, Shan Chang, and Minyi Guo.
\newblock {APR}: Online distant point cloud registration through aggregated
  point cloud reconstruction.
\newblock In \emph{Proceedings of the International Joint Conference on
  Artificial Intelligence}, pages 1204--1212. International Joint Conferences
  on Artificial Intelligence Organization, 2023{\natexlab{a}}.
\newblock Main Track.

\bibitem[Liu et~al.(2023{\natexlab{b}})Liu, Zhu, Zhou, Li, Chang, and
  Guo]{liu2023density}
Quan Liu, Hongzi Zhu, Yunsong Zhou, Hongyang Li, Shan Chang, and Minyi Guo.
\newblock Density-invariant features for distant point cloud registration.
\newblock In \emph{Proceedings of the IEEE/CVF International Conference on
  Computer Vision}, pages 18215--18225, 2023{\natexlab{b}}.

\bibitem[Longuet-Higgins(1981)]{longuet1981computer}
H~Christopher Longuet-Higgins.
\newblock A computer algorithm for reconstructing a scene from two projections.
\newblock \emph{Nature}, 293\penalty0 (5828):\penalty0 133--135, 1981.

\bibitem[Lu et~al.(2021)Lu, Chen, Liu, Zhang, Qu, Liu, and Gu]{lu2021hregnet}
Fan Lu, Guang Chen, Yinlong Liu, Lijun Zhang, Sanqing Qu, Shu Liu, and Rongqi
  Gu.
\newblock {HRegNet}: A hierarchical network for large-scale outdoor {LiDAR}
  point cloud registration.
\newblock In \emph{Proceedings of the IEEE/CVF International Conference on
  Computer Vision}, pages 16014--16023, 2021.

\bibitem[Mei et~al.(2023)Mei, Tang, Huang, Wang, Liu, Zhang, Van~Gool, and
  Wu]{mei2023unsupervised}
Guofeng Mei, Hao Tang, Xiaoshui Huang, Weijie Wang, Juan Liu, Jian Zhang, Luc
  Van~Gool, and Qiang Wu.
\newblock Unsupervised deep probabilistic approach for partial point cloud
  registration.
\newblock In \emph{Proceedings of the IEEE/CVF Conference on Computer Vision
  and Pattern Recognition}, pages 13611--13620, 2023.

\bibitem[Montemerlo et~al.(2002)Montemerlo, Thrun, Koller, Wegbreit,
  et~al.]{montemerlo2002fastslam}
Michael Montemerlo, Sebastian Thrun, Daphne Koller, Ben Wegbreit, et~al.
\newblock {FastSLAM}: A factored solution to the simultaneous localization and
  mapping problem.
\newblock \emph{Association for the Advancement of Artificial Intelligence /
  Innovative Applications of Artificial Intelligence Conference}, 593598, 2002.

\bibitem[Mur-Artal and Tard{\'o}s(2017)]{mur2017orb}
Raul Mur-Artal and Juan~D Tard{\'o}s.
\newblock Orb-slam2: An open-source slam system for monocular, stereo, and
  {RGB-D} cameras.
\newblock \emph{IEEE Transactions on Robotics}, 33\penalty0 (5):\penalty0
  1255--1262, 2017.

\bibitem[Poiesi and Boscaini(2021)]{poiesi2021distinctive}
Fabio Poiesi and Davide Boscaini.
\newblock Distinctive {3D} local deep descriptors.
\newblock In \emph{Proceedings of the International Conference on Pattern
  Recognition}, pages 5720--5727. IEEE, 2021.

\bibitem[Qi et~al.(2017)Qi, Su, Mo, and Guibas]{qi2017pointnet}
Charles~R Qi, Hao Su, Kaichun Mo, and Leonidas~J Guibas.
\newblock {PointNet}: Deep learning on point sets for {3D} classification and
  segmentation.
\newblock In \emph{Proceedings of the IEEE/CVF Conference on Computer Vision
  and Pattern Recognition}, pages 652--660, 2017.

\bibitem[Qin et~al.(2022)Qin, Yu, Wang, Guo, Peng, and Xu]{qin2022geometric}
Zheng Qin, Hao Yu, Changjian Wang, Yulan Guo, Yuxing Peng, and Kai Xu.
\newblock Geometric transformer for fast and robust point cloud registration.
\newblock In \emph{Proceedings of the IEEE/CVF Conference on Computer Vision
  and Pattern Recognition}, pages 11143--11152, 2022.

\bibitem[Qin et~al.(2023)Qin, Chen, Chen, Chen, and Li]{qin2023unifusion}
Zequn Qin, Jingyu Chen, Chao Chen, Xiaozhi Chen, and Xi Li.
\newblock {UniFusion}: Unified multi-view fusion transformer for
  spatial-temporal representation in bird's-eye-view.
\newblock In \emph{Proceedings of the IEEE/CVF International Conference on
  Computer Vision}, pages 8690--8699, 2023.

\bibitem[Ravi et~al.(2020)Ravi, Reizenstein, Novotny, Gordon, Lo, Johnson, and
  Gkioxari]{ravi2020pytorch3d}
Nikhila Ravi, Jeremy Reizenstein, David Novotny, Taylor Gordon, Wan-Yen Lo,
  Justin Johnson, and Georgia Gkioxari.
\newblock Accelerating {3D} deep learning with {PyTorch3D}.
\newblock \emph{arXiv:2007.08501}, 2020.

\bibitem[Rosenfeld and Tsotsos(2019)]{rosenfeld2019intriguing}
Amir Rosenfeld and John~K Tsotsos.
\newblock Intriguing properties of randomly weighted networks: Generalizing
  while learning next to nothing.
\newblock In \emph{Proceedings of the Conference on Computer and Robot Vision},
  pages 9--16. IEEE, 2019.

\bibitem[Rusu et~al.(2009)Rusu, Blodow, and Beetz]{rusu2009fast}
Radu~Bogdan Rusu, Nico Blodow, and Michael Beetz.
\newblock Fast point feature histograms ({FPFH}) for {3D} registration.
\newblock In \emph{Proceedings of the IEEE International Conference on Robotics
  and Automation}, pages 3212--3217. IEEE, 2009.

\bibitem[Simon et~al.(2019)Simon, Amende, Kraus, Honer, Samann, Kaulbersch,
  Milz, and Michael~Gross]{simon2019complexer}
Martin Simon, Karl Amende, Andrea Kraus, Jens Honer, Timo Samann, Hauke
  Kaulbersch, Stefan Milz, and Horst Michael~Gross.
\newblock {Complexer-YOLO}: Real-time {3D} object detection and tracking on
  semantic point clouds.
\newblock In \emph{Proceedings of the IEEE/CVF Conference on Computer Vision
  and Pattern Recognition Workshops}, pages 0--0, 2019.

\bibitem[Sun et~al.(2020)Sun, Kretzschmar, Dotiwalla, Chouard, Patnaik, Tsui,
  Guo, Zhou, Chai, Caine, Vasudevan, Han, Ngiam, Zhao, Timofeev, Ettinger,
  Krivokon, Gao, Joshi, Zhang, Shlens, Chen, and Anguelov]{Sun_2020_CVPR}
Pei Sun, Henrik Kretzschmar, Xerxes Dotiwalla, Aurelien Chouard, Vijaysai
  Patnaik, Paul Tsui, James Guo, Yin Zhou, Yuning Chai, Benjamin Caine, Vijay
  Vasudevan, Wei Han, Jiquan Ngiam, Hang Zhao, Aleksei Timofeev, Scott
  Ettinger, Maxim Krivokon, Amy Gao, Aditya Joshi, Yu Zhang, Jonathon Shlens,
  Zhifeng Chen, and Dragomir Anguelov.
\newblock Scalability in perception for autonomous driving: {Waymo Open
  Dataset}.
\newblock In \emph{Proceedings of the IEEE/CVF Conference on Computer Vision
  and Pattern Recognition (CVPR)}, 2020.

\bibitem[Te et~al.(2018)Te, Hu, Zheng, and Guo]{te2018rgcnn}
Gusi Te, Wei Hu, Amin Zheng, and Zongming Guo.
\newblock {RGCNN}: Regularized graph cnn for point cloud segmentation.
\newblock In \emph{Proceedings of the ACM international conference on
  Multimedia}, pages 746--754, 2018.

\bibitem[Thomas et~al.(2019)Thomas, Qi, Deschaud, Marcotegui, Goulette, and
  Guibas]{thomas2019kpconv}
Hugues Thomas, Charles~R Qi, Jean-Emmanuel Deschaud, Beatriz Marcotegui,
  Fran{\c{c}}ois Goulette, and Leonidas~J Guibas.
\newblock {KPConv}: Flexible and deformable convolution for point clouds.
\newblock In \emph{Proceedings of the IEEE/CVF International Conference on
  Computer Vision}, pages 6411--6420, 2019.

\bibitem[Tombari et~al.(2010)Tombari, Salti, and Di~Stefano]{tombari2010unique}
Federico Tombari, Samuele Salti, and Luigi Di~Stefano.
\newblock Unique signatures of histograms for local surface description.
\newblock In \emph{Proceedings of the European Conference on Computer Vision},
  pages 356--369. Springer, 2010.

\bibitem[Ulyanov et~al.(2018)Ulyanov, Vedaldi, and Lempitsky]{ulyanov2018deep}
Dmitry Ulyanov, Andrea Vedaldi, and Victor Lempitsky.
\newblock Deep image prior.
\newblock In \emph{Proceedings of the IEEE Conference on Computer Vision and
  Pattern Recognition}, pages 9446--9454, 2018.

\bibitem[Wang and Solomon(2019)]{wang2019prnet}
Yue Wang and Justin~M Solomon.
\newblock {PRNet}: Self-supervised learning for partial-to-partial
  registration.
\newblock \emph{Advances in Neural Information Processing Systems}, 32, 2019.

\bibitem[Wu et~al.(2019)Wu, Qi, and Fuxin]{wu2019pointconv}
Wenxuan Wu, Zhongang Qi, and Li Fuxin.
\newblock {PointConv}: Deep convolutional networks on {3D} point clouds.
\newblock In \emph{Proceedings of the IEEE/CVF Conference on Computer Vision
  and Pattern Recognition}, pages 9621--9630, 2019.

\bibitem[Xu et~al.(2020)Xu, Wu, Wang, Zhan, Vajda, Keutzer, and
  Tomizuka]{xu2020squeezesegv3}
Chenfeng Xu, Bichen Wu, Zining Wang, Wei Zhan, Peter Vajda, Kurt Keutzer, and
  Masayoshi Tomizuka.
\newblock {SqueezeSegV3}: Spatially-adaptive convolution for efficient
  point-cloud segmentation.
\newblock In \emph{Proceedings of the European Conference on Computer Vision},
  pages 1--19. Springer, 2020.

\bibitem[Yew and Lee(2018)]{yew20183dfeat}
Zi~Jian Yew and Gim~Hee Lee.
\newblock {3DFeat-Net}: Weakly supervised local {3D} features for point cloud
  registration.
\newblock In \emph{Proceedings of the European Conference on Computer Vision},
  pages 607--623, 2018.

\bibitem[Yew and Lee(2020)]{yew2020rpm}
Zi~Jian Yew and Gim~Hee Lee.
\newblock {RPM-Net}: Robust point matching using learned features.
\newblock In \emph{Proceedings of the IEEE/CVF Conference on Computer Vision
  and Pattern Recognition}, pages 11824--11833, 2020.

\bibitem[Yew and Lee(2022)]{yew2022regtr}
Zi~Jian Yew and Gim~Hee Lee.
\newblock {REGTR}: End-to-end point cloud correspondences with transformers.
\newblock In \emph{Proceedings of the IEEE/CVF Conference on Computer Vision
  and Pattern Recognition}, pages 6677--6686, 2022.

\bibitem[Yu et~al.(2021)Yu, Li, Saleh, Busam, and Ilic]{yu2021cofinet}
Hao Yu, Fu Li, Mahdi Saleh, Benjamin Busam, and Slobodan Ilic.
\newblock {CoFiNet}: Reliable coarse-to-fine correspondences for robust
  pointcloud registration.
\newblock \emph{Advances in Neural Information Processing Systems},
  34:\penalty0 23872--23884, 2021.

\bibitem[Yu et~al.(2022)Yu, Luo, Shu, Huo, Yang, Shi, Guo, Li, Hu, Yuan,
  et~al.]{yu2022dair}
Haibao Yu, Yizhen Luo, Mao Shu, Yiyi Huo, Zebang Yang, Yifeng Shi, Zhenglong
  Guo, Hanyu Li, Xing Hu, Jirui Yuan, et~al.
\newblock {DAIR-V2X}: A large-scale dataset for vehicle-infrastructure
  cooperative {3D} object detection.
\newblock In \emph{Proceedings of the IEEE/CVF Conference on Computer Vision
  and Pattern Recognition}, pages 21361--21370, 2022.

\bibitem[Yu et~al.(2023)Yu, Ren, Zhang, Zhou, Lin, and Dai]{yu2023peal}
Junle Yu, Luwei Ren, Yu Zhang, Wenhui Zhou, Lili Lin, and Guojun Dai.
\newblock {PEAL}: Prior-embedded explicit attention learning for low-overlap
  point cloud registration.
\newblock In \emph{Proceedings of the IEEE/CVF Conference on Computer Vision
  and Pattern Recognition}, pages 17702--17711, 2023.

\bibitem[Zeng et~al.(2017)Zeng, Song, Nie{\ss}ner, Fisher, Xiao, and
  Funkhouser]{zeng20173dmatch}
Andy Zeng, Shuran Song, Matthias Nie{\ss}ner, Matthew Fisher, Jianxiong Xiao,
  and Thomas Funkhouser.
\newblock {3DMatch}: Learning local geometric descriptors from {RGB-D}
  reconstructions.
\newblock In \emph{Proceedings of the IEEE/CVF Conference on Computer Vision
  and Pattern Recognition}, pages 1802--1811, 2017.

\bibitem[Zhang et~al.(2021)Zhang, Zhang, Sun, Zhu, Guo, Qian, and
  Mao]{zhang2021emp}
Xumiao Zhang, Anlan Zhang, Jiachen Sun, Xiao Zhu, Y~Ethan Guo, Feng Qian, and
  Z~Morley Mao.
\newblock {EMP}: Edge-assisted multi-vehicle perception.
\newblock In \emph{Proceedings of the 27th Annual International Conference on
  Mobile Computing and Networking}, pages 545--558, 2021.

\bibitem[Zhang et~al.(2023)Zhang, Yang, Zhang, and Zhang]{zhang20233d}
Xiyu Zhang, Jiaqi Yang, Shikun Zhang, and Yanning Zhang.
\newblock {3D} registration with maximal cliques.
\newblock In \emph{Proceedings of the IEEE/CVF Conference on Computer Vision
  and Pattern Recognition}, pages 17745--17754, 2023.

\bibitem[Zhou et~al.(2016)Zhou, Park, and Koltun]{zhou2016fast}
Qian-Yi Zhou, Jaesik Park, and Vladlen Koltun.
\newblock Fast global registration.
\newblock In \emph{Proceedings of the European Conference on Computer Vision},
  pages 766--782. Springer, 2016.

\bibitem[Zhu et~al.(2022)Zhu, Deng, Zhang, Ji, Mao, Li, and
  Zhang]{zhu2022vpfnet}
Hanqi Zhu, Jiajun Deng, Yu Zhang, Jianmin Ji, Qiuyu Mao, Houqiang Li, and
  Yanyong Zhang.
\newblock {VPFNet}: Improving {3D} object detection with virtual point based
  {LiDAR} and stereo data fusion.
\newblock \emph{IEEE Transactions on Multimedia}, 2022.

\end{thebibliography}
}

\clearpage
\setcounter{page}{1}
\maketitlesupplementary

\section{Detailed Experiment Setup}

\subsection{Comparison Methods}

Considering the lack of genuine unsupervised distant point cloud registration methods at present, we compare EYOC against supervised methods instead. The most compared baselines are the two fully-convolutional methods, FCGF \cite{choy2019fully} and Predator \cite{huang2021predator}. The former utilizes MinkowskiNet for sparse voxel convolution, while the latter builds upon KPConv which classifies as a point convolution method. On the other hand, performances of SpinNet \cite{ao2021spinnet}, D3Feat \cite{bai2020d3feat}, CoFiNet \cite{yu2021cofinet}, and GeoTransformer \cite{qin2022geometric} are quoted verbatim from GCL \cite{liu2023density}.

\subsection{Formal Metric Definition}
\label{sec:appendix_metric}
Given a test set with labels $X_{[d_1,d_2]}=\left\{(S^i,T^i, R^i, t^i)\big|\ ||t^i||_2\in [d_1,d_2]\right\}$ where $S^i,T^i$ are point clouds and $R^i\in SO(3),t^i\in \mathbb{R}^3$ are the ground truth transformation, along with the estimated transformation $\hat{R}^i, \hat{t}^i$, the absolute rotational error and absolute translational error are defined as \cref{eq:re,eq:te}. Please note that we abbreviate $X$ for ${X_{[d_1,d_2]}}$ hereafter to save space where the subscript does not matter.

\begin{equation}
    \mathbf{RE}^{i}_{X} = \arccos{\left(\frac{trace({R^i}^T\hat{R}^i)-1}{2}\right)} \\
    \label{eq:re}
\end{equation}

\begin{equation}
    \mathbf{TE}^{i}_{X} = ||t^i - \hat{t}^i|| \\
    \label{eq:te}
\end{equation}

It is generally observed that, when registration performs well, these errors are usually limited and predictable; However, they could drift randomly during failures, often leading to more than $90\degree$ or $50m$ of error. It is neither interpretable nor repeatable to average the error over all the pairs containing occasional arbitrarily large errors; On the contrary, we often choose to average only those errors of the successful pairs. The registration success is assessed based on the criterion of $S(X,i)=\mathds{1}(RE^{i}_{X}<T_{rot})\times\mathds{1}(TE^{i}_X<T_{trans})$, where $\mathds{1}(\cdot)$ is the Iverson Bracket, and $T_{rot}=5\degree, T_{trans}=2m$ are two generally accepted thresholds. After that, we could calculate the RRE, RTE as the average of RE and TE of succeeded pairs, and RR as the portion of successful pairs over all pairs, as formulated in \cref{eq:rre,eq:rte,eq:rr}:

\begin{equation}
    \mathbf{RRE}_{X} = \frac{1}{\sum\limits_{i=1}^{|X|} S(X,i)}\sum\limits_{i=1}^{|X|} \left(S(X,i) \times \mathbf{RE}^{i}_{X}\right)\\
    \label{eq:rre}
\end{equation}

\begin{equation}
    \mathbf{RTE}_{X} = \frac{1}{\sum\limits_{i=1}^{|X|} S(X,i)}\sum\limits_{i=1}^{|X|} \left(S(X,i) \times \mathbf{TE}^{i}_{X}\right)\\
    \label{eq:rte}
\end{equation}

\begin{equation}
    \mathbf{RR}_{X} = \frac{1}{|X|}\sum\limits_{i=1}^{|X|} S(X,i)\\
    \label{eq:rr}
\end{equation}

Next, mRR is defined as the average of RR over five registration subsets with $||t||\in [d_1,d_2]$ meters,  and the tuple $(d_1,d_2)$ is parameterized according to our specification, \textit{i.e.}, $D_{V2V}=\{(5,10), (10,20), (20,30), (30,40), (40,50)\}$, respectively according to \cref{eq:mrr}:

\begin{equation}
    \mathbf{mRR} = \frac{1}{|D_{V2V}|}\sum\limits_{(d_1,d_2)\in D_{V2V}} \mathbf{RR}_{X_{[d_1,d_2]}}\\
    \label{eq:mrr}
\end{equation}

Lastly, given a dataset $X$ and the estimated correspondences $(j,k)\in C^i$ denoting that $p^j\in S^i, q^k\in T^i$ are a pair of correspondence, the inlier ratio is defined as \cref{eq:ir}:

\begin{equation}
    \mathbf{IR}_{X} = \sum\limits_{i=1}^{|X|} \sum\limits_{(j,k)\in C^i}\frac{\mathds{1}(||R^ip^j+t^i-q^k|| \leq T_{inlier})}{|X|\times |C^i|}\\
    \label{eq:ir}
\end{equation}

Where $T_{inlier}=0.3m$ is the inlier distance threshold.

\section{Method Details}

\begin{figure*}[ht]
    \centering
    \includegraphics[width=0.8\linewidth]{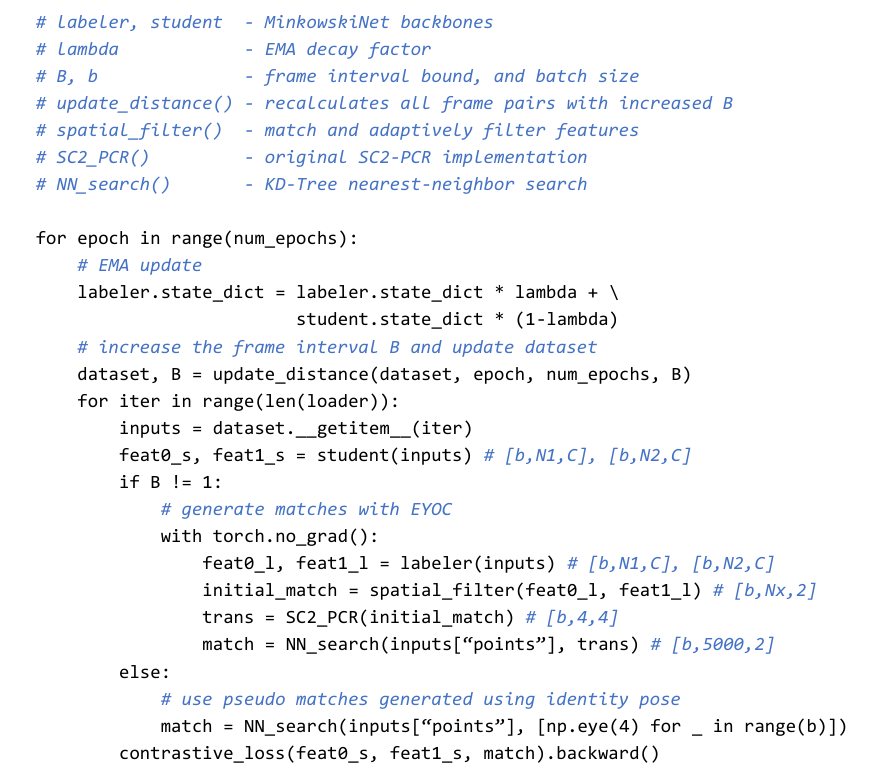}
    \vspace{-0.3cm}
    \caption{Python style pseudo code of the core implementation of EYOC.}
    \label{fig:appendix_pseudo_code}
\end{figure*}

\subsection{Description of SC$^2$-PCR}
We describe the design philosophy and algorithm of SC$^2$-PCR \cite{chen2022sc2} for better stand-alone completeness. SC$^2$-PCR consists of two cascading contributions: a spatial compatibility measure, SC$^2$, and a complete registration pipeline built upon fascinating properties of the SC$^2$ measure.

Past literature have extensively used first order spatial compatibility to measure correspondence quality, which is defined as $M_{x,y}=\left|||p_\mathcal{S}^i-p_\mathcal{S}^j||_2-||p_\mathcal{T}^k-p_\mathcal{T}^l||_2\right|$ for two correspondences $c_x=(p_\mathcal{S}^i, p_\mathcal{T}^k)$ and $c_y=(p_\mathcal{S}^j, p_\mathcal{T}^l)$, where $c_x, c_y\in C$ and $M$ is a matrix of size $|C|\times|C|$. The higher the metric is, the more likely both correspondences $c_x, c_y$ are correct. However, there is still a chance that outliers can be compatible with inliers, making them hard to distinguish. In contrast, the SC$^2$ measure uses $M\cdot M^2$ to measure the number of correspondences in the universe that are simultaneously compatible with two compatible correspondences. As all inliers are compatible with each other, the inliers receive skyrocketing compatibility scores ($\geq \#inliers-2$) and hence are easily identified from outliers.

Built upon the SC$^2$ measure, SC$^2$-PCR takes a two-stage filtering pipeline using the spectral technique to select the most promising seed correspondences and to determine the optimal transformation. The algorithm is both GPU-compatible and non-parametric, resulting in outstanding registration recall, FPS, and generalization capability. All these features entitle SC$^2$-PCR as an ideal labelling algorithm on unlabelled point cloud data.

\subsection{Pseudo Code}
We provide a skeletal structure of EYOC in \cref{fig:appendix_pseudo_code}. All components of EYOC are displayed in the figure. EMA update and distance extension of $B$ precede every epoch, effectively preparing proper weights and data for the next epoch. Inside every training step, if the current frame interval is one, then identity pose will be used for supervised training. Otherwise, the labeler, SR and CR will work together to produce fake correspondence labels. Finally, such labels can be used to calculate a contrastive loss.

\section{Additional Results}

We place the comparison between EYOC and other distant point cloud registration methods, APR and GCL, in \cref{tab:comparison_rebuttal}. While EYOC lags a little bit from the SOTA work GCL with oracle labels on new data (K$\rightarrow$K, N$\rightarrow$N), scoring $-10.2\%$ and $23.8\%$ less mRR on KITTi and nuScenes respectively, EYOC scores consistently better than APR. Moreover, existing supervised methods deteriorate greatly when placed out-of-distribution (K$\rightarrow$W, N$\rightarrow$W), where EYOC gets a lot closer to GCL with $-9.6\%$ and $-2.2\%$ ($\phi\rightarrow$W). When finetuned from the pretrained GCL weights, EYOC achieves even better results with $X\%$ and $Y\%$ gap from GCL instead. We conclude that EYOC, although suffering a performance gap with the SOTA distant PCR method GCL, boasts top-tier performance on unlabelled new data distributions. Furthermore, the defeat can be potentially negated or even overturned should EYOC uses the same group-wise training scheme as GCL, which counts as our future work.

\begin{table}[t]
  \renewcommand{\thetable}{6}
  \centering
  \small
  \resizebox{\linewidth}{!}{
  \begin{tabular}{@{}l|c|c|ccccc@{}}
    \toprule
    \multirow{2}{*}{Method} & Labelled$\rightarrow$ & \multirow{2}{*}{mRR} &\multirow{2}{*}{[5,10]} &\multirow{2}{*}{[10,20]} &\multirow{2}{*}{[20,30]} &\multirow{2}{*}{[30,40]} &\multirow{2}{*}{[40,50]} \\
    &Unlabelled&&&&&&\\
    \midrule
    \multirow{4}{*}{APR} & K$\rightarrow$ K         &77.9 	&99.2 	&96.8 	&88.3 	&67.6 	&37.8\\ 
    & K$\rightarrow$ W                              &69.1 	&97.1 	&87.4 	&68.2 	&53.2 	&39.8\\ 
    & N$\rightarrow$ N                              &58.8 	&99.5 	&85.6 	&43.8 	&45.7 	&19.2\\ 
    & N$\rightarrow$ W                              &68.4 	&95.2 	&84.5 	&60.0 	&56.1 	&46.3\\ 

    \midrule
    \multirow{4}{*}{GCL} & K$\rightarrow$ K         &93.5 	&99.0 	&98.8 	&96.1 	&91.7 	&82.0\\ 
    & K$\rightarrow$ W                              &88.0 	&100.0 	&99.0 	&91.8 	&79.9 	&69.1\\
    & N$\rightarrow$ N                              &85.5 	&99.3 	&97.7 	&91.8 	&77.8 	&60.7\\ 
    & N$\rightarrow$ W                              &80.6 	&99.0 	&95.2 	&81.2 	&67.6 	&60.2\\ 

    \midrule
    \multirow{3}{*}{EYOC} & $\phi\rightarrow$ K     &83.2 	&99.5 	&96.6 	&89.1 	&78.6 	&52.3 \\
    & $\phi\rightarrow$ W                           &78.4 	&97.6 	&91.3 	&78.2 	&65.5 	&59.3 \\
    & $\phi\rightarrow$ N                           &61.7 	&96.7 	&85.6 	&61.8 	&37.5 	&26.9 \\
    \bottomrule
  \end{tabular}
  }
  \vspace{-0.3cm}
  \caption{\textbf{Comparison of EYOC, FCGF+APR(a) and GCL+Conv}, where \textit{K, W, N, $\phi$} represent KITTI, WOD, nuScenes and scratch. While we observe GCL $>$ EYOC $>$ APR in supervised settings, EYOC excels on new unlabelled data by unsupervised finetuning. This will be included in the revision.}
  \vspace{-0.2cm}
  \label{tab:comparison_rebuttal}
\end{table}

\section{Discussions}

\paragraph{Compatibility with previous literature.} Moreover, we notice that Hypothesis \ref{hyp:corr} would hint that point cloud features would deteriorate (\textit{i.e.}, move) on the feature space slower than linear functions relative to the distance-to-LiDAR (\textit{e.g.}, radical functions). We argue that this does not contradict previous literature \cite{liu2023density} which found the relation to be linear; While previous literature looked into the in-domain performance of converged models, we are looking into the out-of-domain performance of models during training. It is natural for networks to behave differently on seen and unseen data.

\paragraph{Performance Upper Bound.} We note that better network weight boosts SC$^2$-PCR’s label quality and better labels promote network performance. Consequently, EYOC’s upper bound should be the combination of \textit{(i) bound of SC$^2$-PCR labels given a hypothetical oracle feature extractor}, and \textit{(ii) bound of a feature extraction network given an oracle labeler algorithm, i.e., supervised training}. Our inclination is that bound\textit{ (ii)} is tighter and contributes a major decrease in the upper bound while SC$^2$-PCR, \textit{i.e.}, bound \textit{(i)}, plays a minor part, as evidenced by the RR@[40m,50m] values consistently remaining below 65\%, far from the $90$+ RR reported in SC$^2$-PCR.

\paragraph{Error Accumulation.} We believe EYOC is capable of avoiding error accumulation thanks to the induction bias present in the filtering pipeline. Pose estimators such as SC$^2$-PCR tend to output poses that are either close to perfect (\cref{fig:visualization}) or randomly distributed in the SO(3) space. While the presence of suboptimal features may decrease the percentage of perfect poses, they do not incur significant errors on all output poses, and the precise poses stay correct. In return, during instances of failure, the random erroneous positives and negatives are scattered in feature space (as anything could be matched with anything else), effectively canceling each other out, yielding limited impact compared to the correct labels.

\begin{figure}[t]
  \centering
%   \vspace{-0.2cm}
%   \includegraphics[width=\linewidth]{figure/analysis.pdf}
  \includegraphics[width=\linewidth]{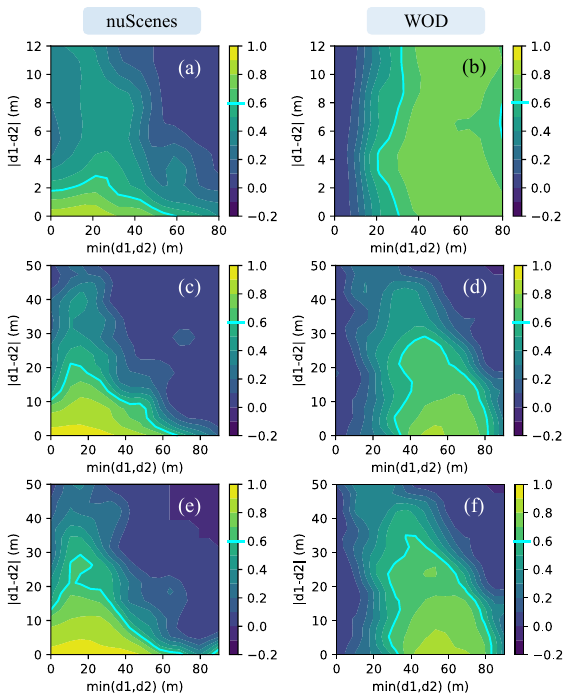}
  \vspace{-0.7cm}
  \caption{\textbf{Visual groundings} for our hypothesis on (a,c,e) nuScenes \cite{Caesar_2020_CVPR} and (b,d,f) WOD \cite{Sun_2020_CVPR}. Cosine similarity of correspondences with its distance to two LiDARs, $d_1, d_2$, is displayed for $I\in [1,1]$ (top), $I\in [1,15]$ (middle), and $I\in [1,30]$ (bottom). Decision boundaries at $s_{thresh}=0.6$ are highlighted in cyan.}
  \label{fig:dist_sim_nus_wod}
  \vspace{-0.2cm}
\end{figure}

% \subsection{More on Time Analysis}
% \label{sec:appendix_time_analysis}
% We further explain why EYOC has a lower training time than its supervised counterpart, FCGF \cite{choy2019fully}. As pointed out in \cref{sec:time}, threaded dataloaders are not allowed to use GPU as specified in Pytorch, but we do notice that canceling the threading enables GPUs in dataloaders, meaning that we actually can perform NN-search on GPU for FCGF. However, this cancels the boost caused by pipelining multiple dataloader threads with the trainer thread, resulting in a surge of data loading time to $1885$ms instead. The only reasonable choice is to make drastic changes to the original code, moving the NN-search step to the trainer thread completely with the help of Pytorch3D, which would result in a time consumption similar to that of EYOC.

\section{Visualization}

\subsection{Spatial Filtering on other Datasets}
\label{sec:appendix_other_datasets}
We display the spatial feature similarity results on WOD and nuScenes in \cref{fig:dist_sim_nus_wod}, where $d_1, d_2$ denotes the distance from a correspondence to the two LiDAR centers, and the similarity is indicated by brightness. The decision boundary of $s_thresh=0.6$ is highlighted in cyan, similar to \cref{fig:theoretical_similarity}. WOD exhibits almost identical traits to those on KITTI, showing a drastic feature deterioration in the close-to-LiDAR regions as well as the extremely far regions, and cutting off at 40m would almost always cut the closer half below 0.6 similarity, indicating the similarity between the two filtering strategies. On the other hand, nuScenes displays a similar pattern where high-similarity regions are clustered 20 meters away from the LiDAR. Compared to those on KITTI or WOD, the pinnacle region in nuScenes is slightly shifted towards the LiDAR compared with the other two datasets, due to the lower LiDAR resolution and consequently lower density. In nuScenes, it would be improper to cut off at 40m, although the training does converge and has decent performance as reported in \cref{tab:comparison}. While this phenomenon is attributed to the discrepancy between KITTI-style datasets and nuScenes-style datasets, we also highlight that EYOC is robust under such discrepancies even when the patterns for the pretraining dataset (WOD) largely differ from the actual one on the finetuning dataset (nuScenes).

\subsection{Registration Results}
\begin{figure*}[t]
  \centering
  \includegraphics[width=0.45\linewidth]{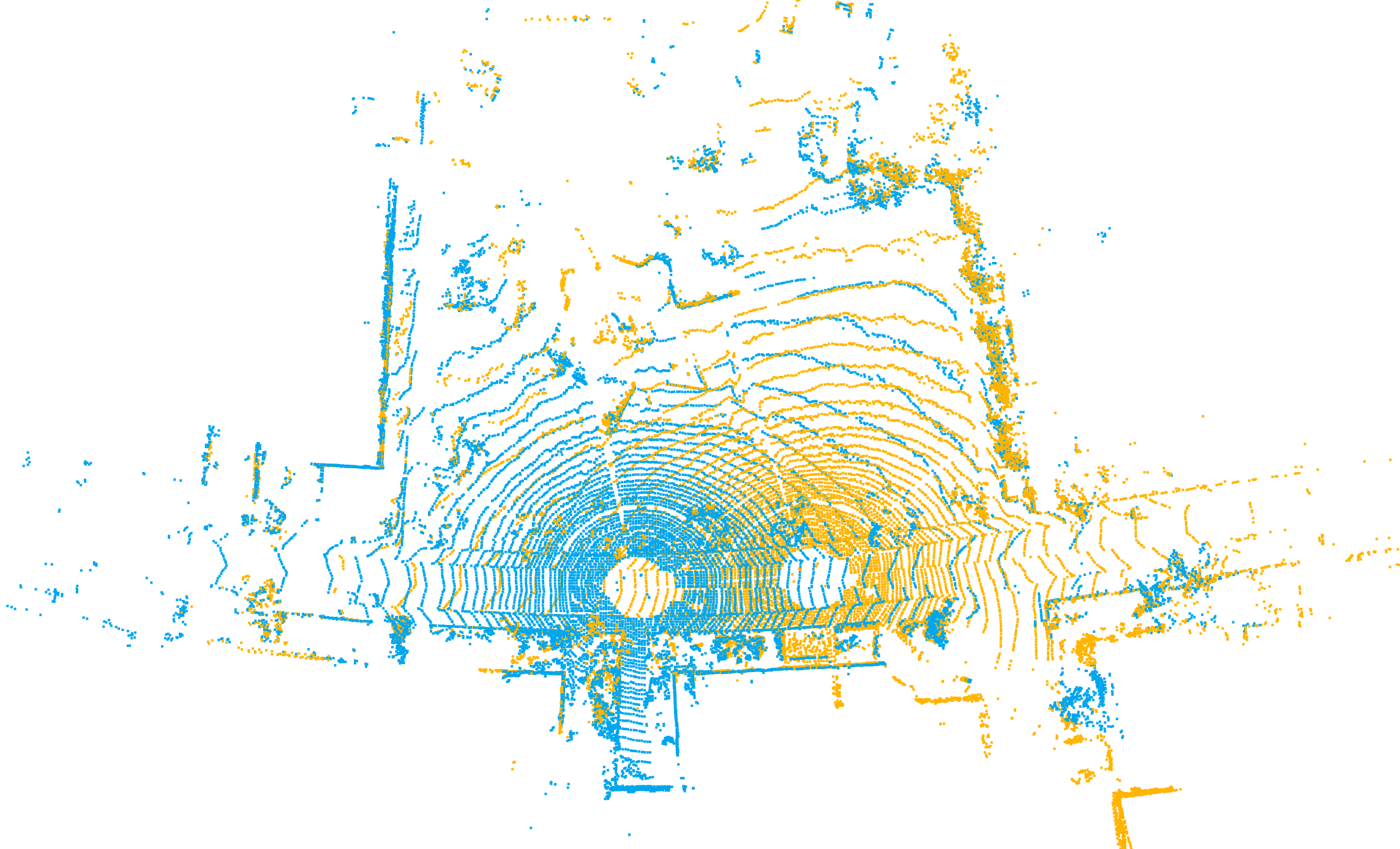} \hfill
  \includegraphics[width=0.45\linewidth]{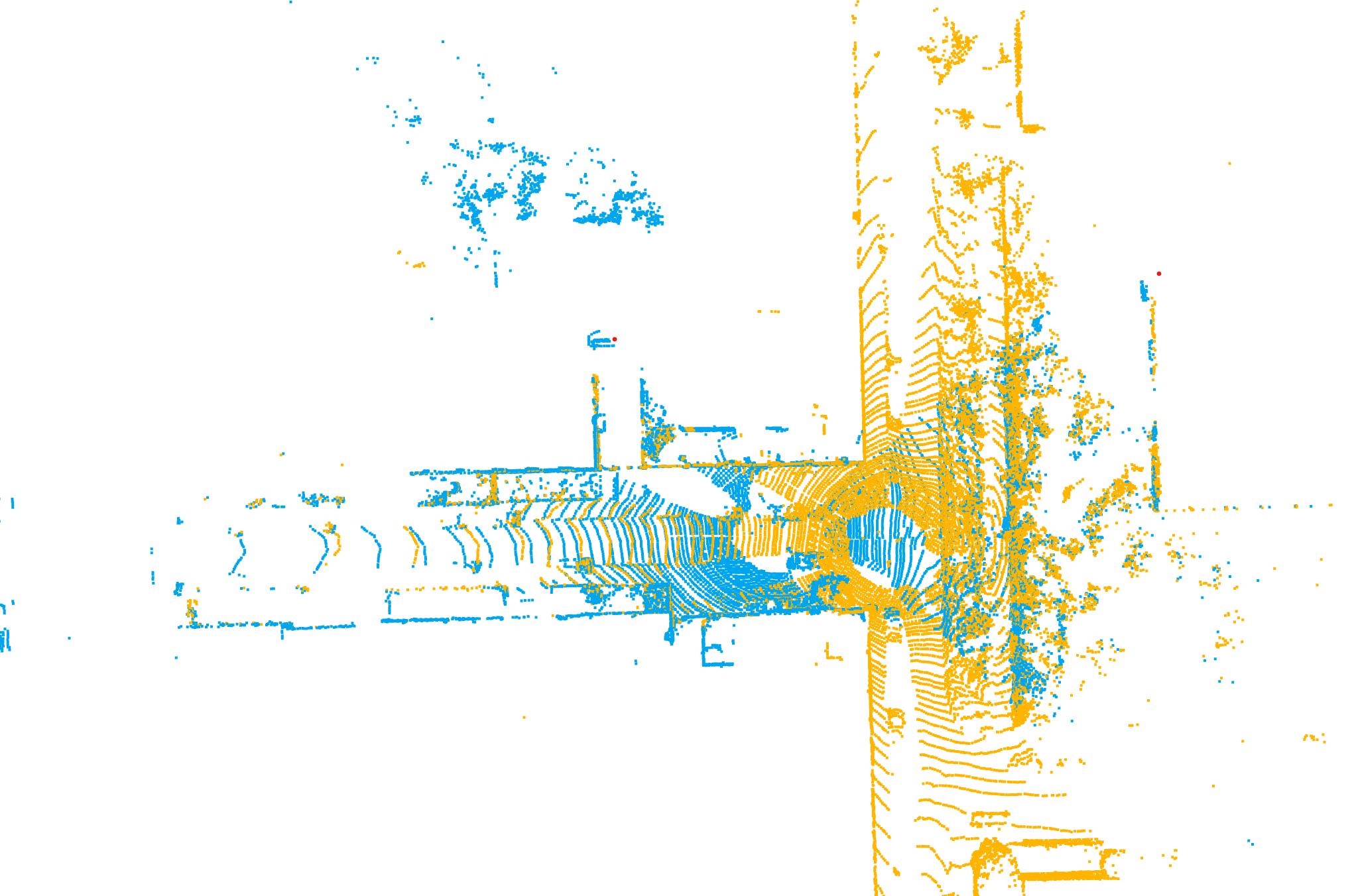} \hfill
  \caption{Registration results of EYOC on KITTI \cite{Geiger2012CVPR}.}
  \label{fig:reg_results_kitti}
%   \vspace{-0.2cm}
\end{figure*}

\begin{figure*}[t]
  \centering
  \includegraphics[width=0.45\linewidth]{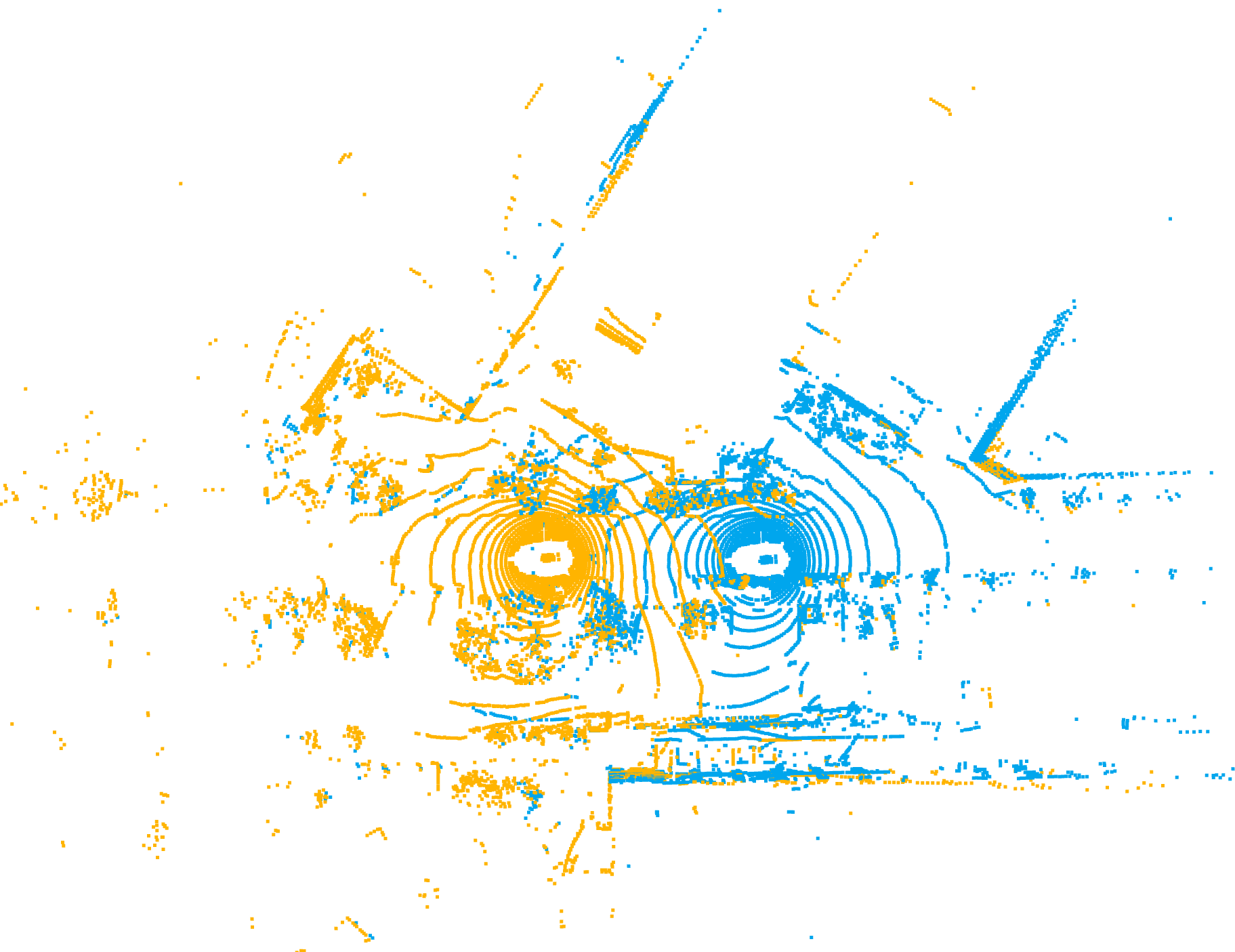} \hfill
  \includegraphics[width=0.45\linewidth]{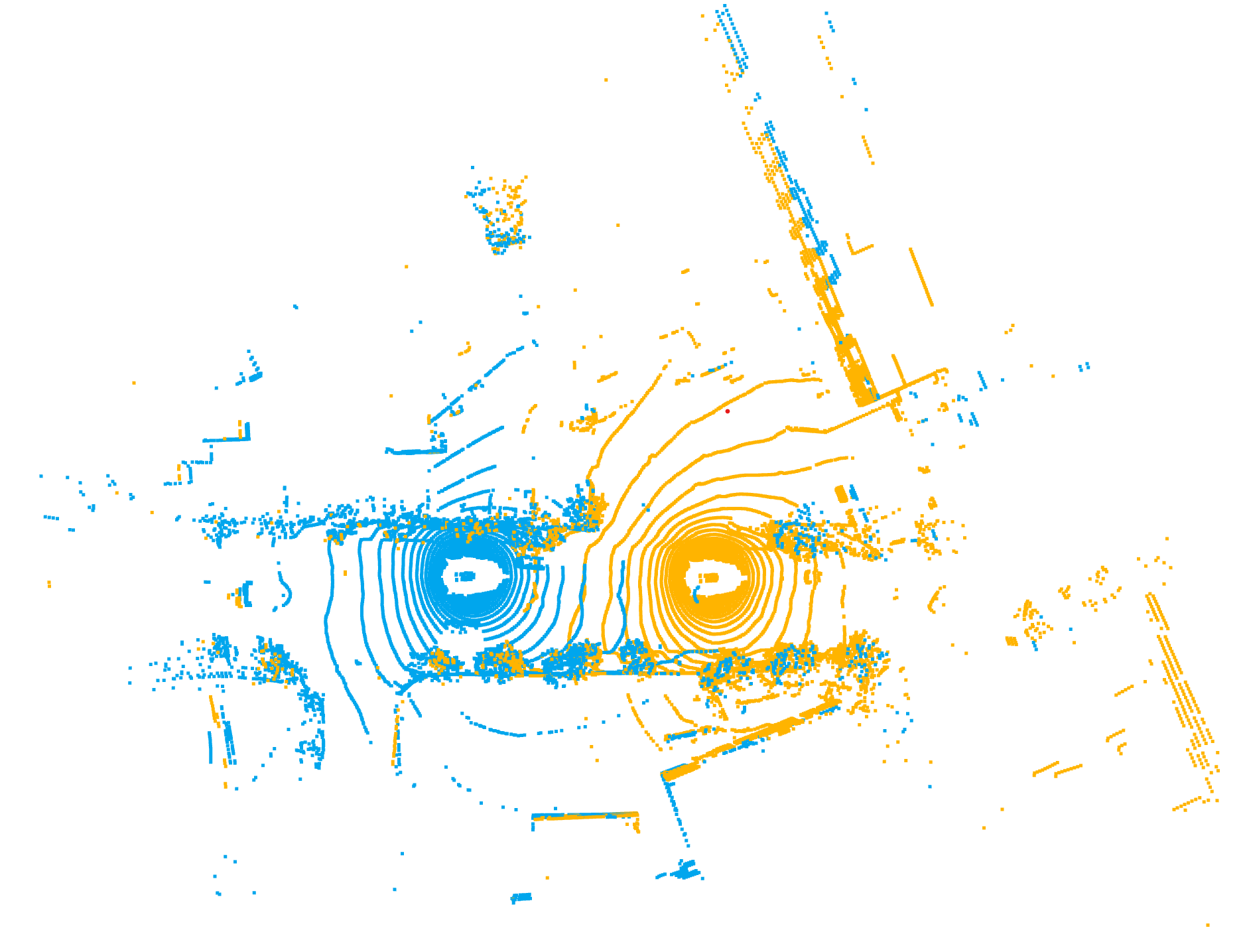} \hfill
  \caption{Registration results of EYOC on nuScenes \cite{Caesar_2020_CVPR}.}
  \label{fig:reg_results_nuscenes}
%   \vspace{-0.2cm}
\end{figure*}

\begin{figure*}[t]
  \centering
  \includegraphics[width=0.5\linewidth]{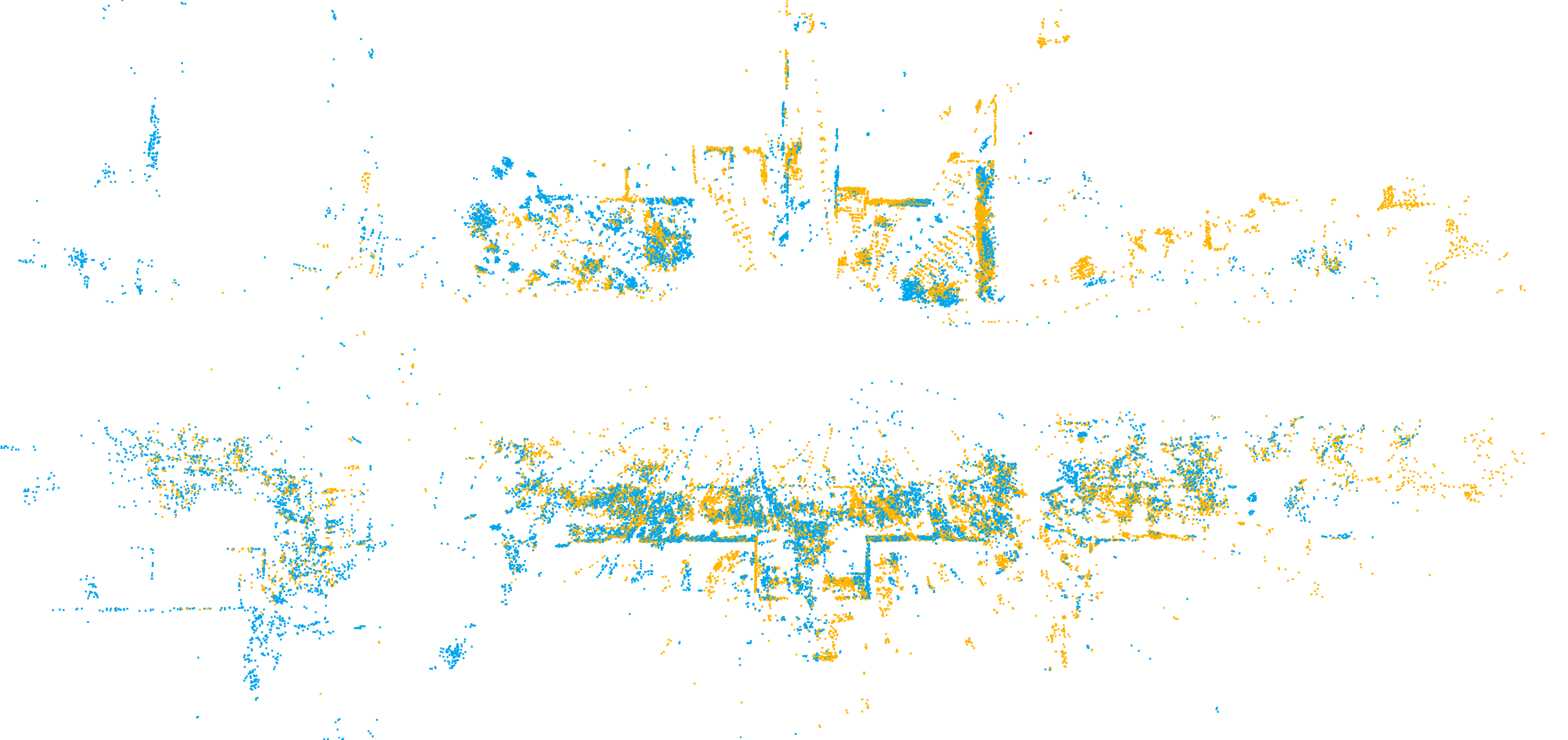} \hfill
  \includegraphics[width=0.45\linewidth]{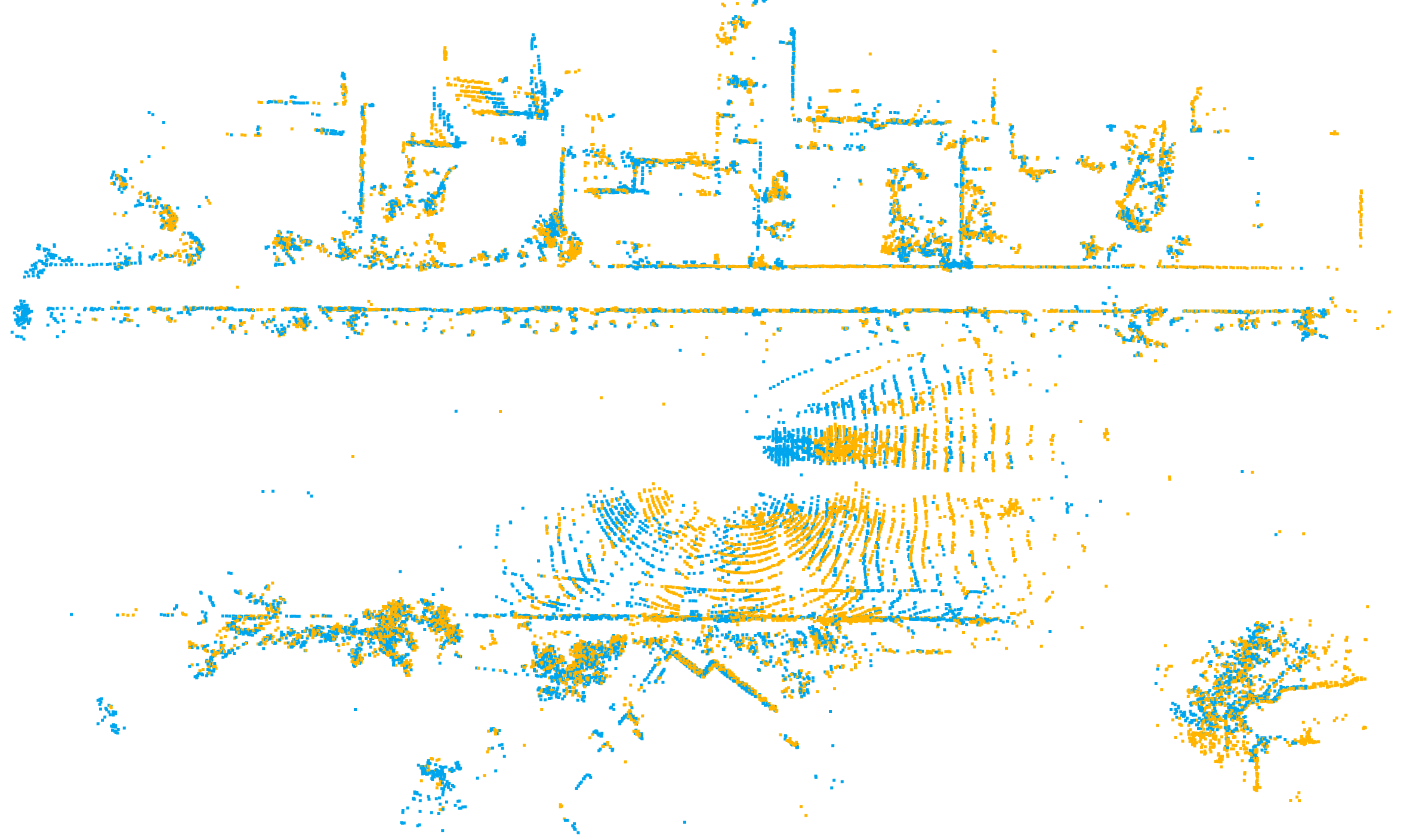} \hfill
  \caption{Registration results of EYOC on WOD \cite{Sun_2020_CVPR}, demonstrated using only the second return.}
  \label{fig:reg_results_waymo}
%   \vspace{-0.2cm}
\end{figure*}

We display the registration results of EYOC on KITTI, nuScenes and WOD in \cref{fig:reg_results_kitti,fig:reg_results_nuscenes,fig:reg_results_waymo}.

\end{document}